\documentclass[final,3p,times]{elsarticle}

\usepackage{amsmath}
\usepackage{amssymb}
\usepackage{graphicx}
\usepackage{verbatim}
\usepackage{url}
\usepackage{subfig}

\def\x{{\mathbf x}}

\newcommand{\etal}[0]{\emph{et. al. }}
\newcommand{\norm}[1]{\left\Vert#1\right\Vert}
\newcommand{\abs}[1]{\left\vert#1\right\vert}

\newcommand\vect[1]{{\bf#1}}
\newcommand\matr[1]{{\bf#1}}
\newcommand\alphabf{{\boldsymbol{\alpha}}}

\DeclareMathOperator{\trace}{trace}
\DeclareMathOperator{\MSE}{MSE}

\DeclareMathOperator{\GCV}{GCV}

\journal{Applied and Computational Harmonic Analysis}

\begin{document}

\begin{frontmatter}

\title{The Projected GSURE for
Automatic Parameter Tuning in Iterative Shrinkage Methods}

\author[csa]{R.~Giryes\corref{cor1}}
\ead{raja@cs.technion.ac.il}
\author[csa]{M.~Elad}
\ead{elad@cs.technion.ac.il}
\author[eea]{Y.~C.~Eldar}
\ead{yonina@ee.technion.ac.il}

\cortext[cor1]{Corresponding author}

\address[csa]{The Department of Computer Science,
        Technion -- Israel Institute of Technology,
        Haifa 32000, Israel}
\address[eea]{The Department of Electrical Engineering,
        Technion -- Israel Institute of Technology,
        Haifa 32000, Israel}

\begin{abstract}
Linear inverse problems are very common in signal and image
processing. Many algorithms that aim at solving such problems
include unknown parameters that need tuning. In this work we focus
on optimally selecting such parameters in iterative shrinkage
methods for image deblurring and image zooming. Our work uses the
projected Generalized Stein Unbiased Risk Estimator (GSURE) for
determining the threshold value $\lambda$ and the iterations
number $K$ in these algorithms. The proposed parameter selection
is shown to handle any degradation operator, including ill-posed
and even rectangular ones. This is achieved by using GSURE on the
projected expected error. We further propose an efficient greedy
parameter setting scheme, that tunes the parameter while iterating
without impairing the resulting deblurring performance. Finally,
we provide extensive comparisons to conventional methods for
parameter selection, showing the superiority of the use of the
projected GSURE.
\end{abstract}


\end{frontmatter}


\section{Introduction}
\label{sect:intro}

In many applications in  signal and image processing there is a
need for solving a linear inverse problem
\cite{Vogel2002ComputMethods,Bertero98IntroducInverse}. Here we
consider the scenario in which an original image $\vect{x}$ is
degraded by a linear operator $\matr{H}$ followed by an additive
white Gaussian noise $\vect{w}$. Our observation can be written as
\begin{equation}\label{eq:blur}
\vect{y} = \matr{H}\vect{x} + \vect{w},
\end{equation}
where the goal is to reconstruct the original image $\vect{x}$
from $\vect{y}$. In our setting, there is no Bayesian prior on
$\vect{x}$; it is therefore treated as a deterministic unknown.

There are many techniques in the  literature for this task that
focus on different objectives
\cite{Vogel2002ComputMethods,Bertero98IntroducInverse,Takeda08dDeblurring,
foi06Shape--adaptive,Dabov07imagedenoising,Fadili07Sparse--Representation,
Daubechies04Iter--threshold,Elad07awide-angle,Figueiredo05BOA,BE07a,KE98}.
Most of these methods include parameters that require tuning. Some
of the algorithms are iterative, and thus the number of iterations
needs to be set as well. Tuning of such parameters is generally
not a trivial task. Very often, the objective in these problems is
recovery of the signal with a minimal Mean-Squared Error (MSE)
between the original image and the estimated one
\cite{eldar2009biased}. In this case, ideally, the parameters
should be chosen such that this MSE is minimized. However, since
the original image $\vect{x}$ is deterministic, the MSE  will
depend in general on the unknown $\vect{x}$. Consequently, we
cannot know what choice of parameters minimizes the true MSE. In
practice, therefore, a common approach to parameter tuning is
still manual, by looking at the reconstructed result and judging
its quality subjectively.

The literature offers several automatic  ways for choosing the
parameters of the reconstruction algorithm
\cite{Vogel2002ComputMethods}. One such method is the Unbiased
Predictive Risk Estimator (UPRE) \cite{Mallows73Cp}, which is valid
only for the case that the reconstruction operator is linear. Other
popular tools are the generalized cross validation (GCV) technique
with its many variants \cite{Golub1979Generalized}, and the L-Curve
method \cite{Hansen93L-curve}, both available for general
reconstruction algorithms. The number of iterations can be chosen
using the discrepancy principle \cite{Morozov66OnSol}, such that the
norm of the error between the degraded image and reconstructed image
is close to the noise variance.

An alternative technique for parameter tuning is the Stein Unbiased
Risk Estimator (SURE)
\cite{Stein73Estimate--mean,Stein81Estimate--mean}. SURE provides an
unbiased estimate of the MSE for a candidate solver, including
non-linear ones. Minimizing this estimated MSE, leads to the
automatic tuning desired. The original SURE by Stein is limited to
the case of denoising of white additive Gaussian noise. One of the
best known algorithms that uses SURE is Donoho's SureShrink
denoising algorithm \cite{Donoho95Adapting-wavelet}. There are also
other contributions along these lines, as in
\cite{Blu07SURE-LET,Pesquet97adaptive--signal,Benazza-Benyahia05Building--robust}.

Recently, a generalization of SURE (GSURE) \cite{Eldar09GSURE} has
been developed for more general models that can be described in
the form of exponential family distribution. This provides a new
opportunity for choosing  the parameters automatically in a much
more diverse set of inverse problems. In the case where $\matr{H}$
is rank-deficient, or even rectangular (with more columns than
rows), the inverse problem becomes ill-posed. In such a scenario,
GSURE can at best estimate the MSE in the range of $\matr{H}^T$.
We refer hereafter to a tuning of the algorithm's parameters with
this estimation as the projected GSURE.

The reconstruction algorithms used in our work belong to the {\em
iterated shrinkage} family of methods
\cite{Daubechies04Iter--threshold,Elad07awide-angle,Elad07Coord--subspace,Figueiredo03anem},
In these algorithms, the image is known to have a sparse
representation over a dictionary. Reconstruction is obtained by an
iterative process composed of applying the dictionary and the
degradation operator with their conjugates, in addition to a
point-wise shrinkage operation. In these algorithms the threshold
value $\lambda$ in the shrinkage parameter needs to be tuned,
along with the number of iterations $K$ to be performed.

The work reported in \cite{Vonesch08GSURE} develops a special case
of GSURE for the case where the blur operator $\matr{H}$ is
assumed to be full-rank. This work also addresses automatic tuning
of the parameters $\lambda$ and $K$ in an iterated shrinkage
algorithm, used for the deblurring problem. For a pre-chosen
number of iterations, the GSURE is used to select a fixed value of
$\lambda$ that optimizes the overall estimated MSE. For a given
$\lambda$, the GSURE is again used to determine the number of
iterations. We refer to this method of setting $\lambda$ and $K$
as the {\em global method}.

Similar to the work in \cite{Vonesch08GSURE}, this paper proposes
to tune $\lambda$ and $K$ for iterated shrinkage algorithms, which
are geared towards solving inverse problems in image processing.
Our work extends the applicability of GSURE to ill-posed inverse
problems, using the projected GSURE. We show that turning from the
plain GSURE to its projected version, the performance of the
parameter tuning improves substantially.

In addition, we propose an alternative to the global method
mentioned above, where the value of $\lambda$ is chosen by
minimizing the estimated MSE at each iteration, thereby obtaining
a different value of $\lambda$ per iteration. We refer to this
technique as the {\em greedy method}. The main benefit of such a
strategy is the natural way with which the number of iterations is
set simultaneously -- the iterations can be stopped when the
estimated MSE improvement is under a certain threshold. In order
to further improve the performance with regard to the overall
estimated MSE in this greedy approach, $\lambda$ in each iteration
can be chosen with a look-ahead on the estimated MSE in the next
few iterations.

This paper is organized as follows: In  Section
\ref{sect:iterative_shrinkage} we present the deblurring and image
scaling (zooming) problems, and the iterative shrinkage algorithms
used for solving them.
Section~\ref{sect:ConvParametrTuningMethods} describes several
conventional techniques for parameter setting (GCV, L-curve, and
the discrepancy method). Section \ref{sect:sure_gsure} presents
the SURE with its generalization, as developed in
\cite{Eldar09GSURE}. In Section \ref{sect:param_set_gsure} we use
GSURE  for the task of parameter selection in the iterative
shrinkage algorithms using the two approaches -- global and
greedy. The results of these methods are presented in Section
\ref{sect:results}.
We conclude in Section \ref{sect:conclusion} with a summary of
this work, and a discussion on future research directions.


\section{Iterative Shrinkage Methods}
\label{sect:iterative_shrinkage}

To define a linear inverse problems, we need to specify the model
of the original image, the noise characteristics, and the
degradation operator. We consider two types of operators
$\matr{H}$. The first is a blur operator, and the second is a
scale-down operator that defines a scale-up problem. In the blur
case, $\matr{H}$ is a square matrix that blurs each pixel by
replacing it with a weighted average of its neighbors. In the
scale-down scenario, $\matr{H}$ is composed of two operators
applied sequentially, the first blurs the image and the second
down-samples it. In both settings we assume that the operators are
known, and the additive noise $\vect{w}$ is white Gaussian with
known variance $\sigma^2$. Our goal in both settings is to
reconstruct the original image $\vect{x}$ without assuming any
Bayesian prior.

In order to evaluate the quality of the reconstruction, we measure
the MSE between the reconstructed image and the original one,
\begin{equation}
\label{eq:MSE}
\MSE(\vect{\hat{x}})=E\left[\norm{\vect{x}-\hat{\vect{x}}}_2^2\right].
\end{equation}
This criterion is very commonly used in image processing despite
its known shortcomings \cite{Bovik}. Two related measures that are
also common are the Peak Signal to Noise Ratio (PSNR), and the
Improvement in Signal to Noise Ratio (ISNR)\footnote{These are
defined as $PSNR=10\log_{10}\left(255^2/MSE\right)$ and
$ISNR=10\log_{10}\left(MSE_{ref}/MSE\right)$, where $MSE_{ref}$ is
a reference-MSE to compare against.}, which are both inversely
proportional to the MSE. Thus, all of the criteria favor low MSE.

Since the goodness of our fit is measured by the MSE, a natural
approach would be to choose an estimate $\hat{\vect{x}}$ that
directly minimizes the MSE. Unfortunately, however, the MSE
depends in general on $\vect{x}$ and therefore cannot be minimized
directly. This difficulty is explained in more detail in
\cite{eldar2009biased}. To avoid direct MSE minimization, in the
context of imaging it is often assumed that $\vect{x}$ has a
sparse representation $\alphabf$ under a given dictionary
$\matr{D}$ \cite{Elad09Review}. This knowledge of $\vect{x}$ is
then used together with a standard least-squares criterion to find
an estimate $\hat{\vect{x}}$. More specifically, a popular
approach to recover $\vect{x}$ is to first seek the value of
$\alphabf$ that minimizes an objective of the form:
\begin{equation}
\label{eq:deblur_obj} f(\alphabf) =
\frac{1}{2}\norm{\vect{y}-\matr{HD}\alphabf}_2^2 + \lambda
\rho(\alphabf).
\end{equation}
The first term requires that the reconstructed image, after
degradation, should be close to the measured image $\vect{y}$. The
second term forces sparsity using the penalty function $\rho$. The
minimizer of this function is the representation of the desired
image, and its multiplication by $\matr{D}$ gives the
reconstructed image. Although intuitive, this strategy does not in
general minimize the MSE.

The minimization of $f(\alphabf)$ in (\ref{eq:deblur_obj}) is
typically performed using an iterative procedure. An appealing
choice is the iterative shrinkage methods
\cite{Daubechies04Iter--threshold,Elad07awide-angle}. These
algorithms are iterative, applying the matrices $\matr{H}$ and
$\matr{D}$ and their adjoints, and an additional element-wise
thresholding operation in each iteration. Consequently, these
methods are well-suited for high-dimensional problems with sparse
unknowns. This family of techniques includes the Separable
Surrogate functionals (SSF) method
\cite{Daubechies04Iter--threshold}, the Parallel Coordinate
Descent (PCD) method \cite{Elad06PCD,Elad07Coord--subspace}, and
some others. In this work we focus on the SSF and PCD techniques.
The SSF algorithm proposes the iteration formula
\begin{equation} \label{eq:ssf_iter}
\alphabf_{k+1} =
\mathcal{S}_{\rho,\lambda/c}\left(\frac{1}{c}(\matr{HD})^T(\vect{y}-
\matr{HD}\alphabf_k)+\alphabf_k\right),
\end{equation}
which is proven to converge \cite{Daubechies04Iter--threshold} to
a local minimum of (\ref{eq:deblur_obj}). In the case where $\rho$
is convex, this is also the global minimum. The term
$\mathcal{S}_{\rho,\lambda/c}$ is a shrinkage operator dependant
on $\rho$ with threshold $\lambda/c$. The constant $c$ depends on
the combined operator $\matr{HD}$. The resulting estimator is
\begin{equation}
\label{eq:estimator} \vect{\hat{x}} = h_{\lambda,K}(\vect{y}) =
\matr{D}\alphabf_{K},
\end{equation}
where $K$ is the number of SSF iterations performed. When our task
is minimizing the objective in (\ref{eq:deblur_obj}), $K$ is
chosen to be the point that the decreasing rate is under a certain
threshold. When aiming at minimization of the MSE, both $\lambda$
and $K$ need to be tuned carefully: in many cases, from a certain
iteration, the objective value continues to descend, while the MSE
value increases.

In \cite{Elad06PCD,Elad07Coord--subspace} an alternative iterated
shrinkage approach is presented, which leads to faster
convergence. In this technique, termed PCD, the iteration formula
is given by
\begin{equation}
\label{eq:PCD_solution} \alphabf_{k+1}  =  \alphabf_k -
\mu(\vect{v}_k - \alphabf_k)
\end{equation}
where
\begin{equation}
\vect{v}_k = \mathcal{S}_{\rho,\matr{W} \lambda}\left(
\matr{W} (\matr{HD})^T(\vect{y} - \matr{HD}\alphabf_k
) + \alphabf_k \right).
\end{equation}
The diagonal matrix $\matr{W}$ contains the norms of the columns
of $\matr{HD}$, which are computed off line. Once $\vect{v}_k$ is
computed, a line search is performed along the trajectory
$\vect{v}_k - \alphabf_k$, using the parameter $\mu$. This method
is also proven to converge \cite{Elad07Coord--subspace}. Our
estimator in this case is computed by (\ref{eq:estimator}) like in
the SSF.

In order to use the SSF and the PCD methods for the inverse problems
mentioned, the penalty function $\rho$ and the dictionary $\matr{D}$
need to be chosen as well as the $\lambda$ and the number of
iterations. Following Elad \etal  \cite{Elad07Coord--subspace}, we use the
function
\begin{eqnarray}
\label{eq:smooth_l_p_norm} \rho_s(\alpha) = \abs{\alpha} -
s\log\left(1+\frac{\abs{\alpha}}{s}\right) & s \in (0,\infty).
\end{eqnarray}
This is a smoothed version of the $\ell_p$-norm in the  range
$1<p<2$. The parameter $s$ defines which $\ell_p$-norm we are
approximating in this range. For small values of $s$ (e.g.,
$s=0.001$) this function serves as a smoothed version of the
$\ell_1$-norm, and we use it as the penalty function throughout this
work. This choice of $\rho$ dictates the shape of the shrinkage
operator $\mathcal{S}$ \cite{Elad07Coord--subspace}.

There are various possibilities for  choosing $\matr{D}$. In our
simulations, and following the work reported in
\cite{Figueiredo05BOA,Figueiredo03anem,Elad07awide-angle}, we use
the undecimated Haar wavelet with 3 layers of resolution. The
motivation for this choice is that when looking for sparse
representations, redundant dictionaries are favorable.

The parameter $\lambda$ which determines the weight of the penalty
function also needs setting, as do the number of iterations.
Ideally, we would like to choose the parameters that minimize the
MSE between the original image $\vect{x}$ and its estimate.
However, as we have seen already, the dependence of the MSE on
$\vect{x}$ precludes its direct minimization. We now turn to
present several methods that aim to bypass this difficulty.


\section{Conventional Parameter Tuning methods}
\label{sect:ConvParametrTuningMethods}

The main approaches the literature offers for automatic parameter
setting are the GCV \cite{Golub1979Generalized}, the L-Curve
\cite{Hansen93L-curve} and the discrepancy principle
\cite{Morozov66OnSol}. For linear deconvolution methods, the
parameter selection can also be based on the Unbiased Predictive
Risk Estimator (UPRE) \cite{Mallows73Cp}; however, the algorithms
we treat here are nonlinear.

In this section we accompany the description of these methods with
an image deblurring experiment along the following lines: Working
on the image {\tt cameraman}, we use the blur kernel
\begin{eqnarray}
\label{eq:ker1} ker1: & h(x_1,x_2)=\frac{1}{1+x_1^2+x_2^2}, &
x_1,x_2=-7,...,7
\end{eqnarray}
and an additive white Gaussian noise with variance $\sigma^2= 2$.
This experiment and its chosen parameters are adopted from Figueiredo \etal \cite{Figueiredo05BOA}.

\subsection{The Generalized Cross Validation (GCV) Method}
\label{sect:GCV}

For a reconstruction algorithm $h_{\lambda,K}(\cdot)$ and a
degraded image $\vect{y}$, we define the residual of the algorithm
to be
\begin{eqnarray}
\label{eq:reconst_resid}
\vect{r_{h_{\lambda,K}(\vect{y})}} = \matr{H}h_{\lambda,K}(\vect{y}) - \vect{y}.
\end{eqnarray}
In the case where $h_{\lambda,K}(\cdot)$ is linear, the GCV method
selects the parameters to be those that minimize the GCV functional
given by
\begin{eqnarray}
\label{eq:GCV_fun}
\GCV({\lambda,K}) = \frac{\frac{1}{n_\vect{y}}
\norm{\vect{r}_{h_{\lambda,K}(\vect{y})}}_2^2}{\left[\frac{1}{n_\vect{y}}\trace(\matr{I}-
\matr{H}h_{\lambda,K}(\cdot))\right]^2}&,
\end{eqnarray}
where $\matr{I}$ is the identity matrix, and $n_\vect{y}$ is the
length of $\vect{y}$. In high dimensions, the trace in the
denominator of (\ref{eq:GCV_fun}) is costly to compute directly.
Stochastic estimation of this expression can be used instead
\cite{Hutchinson90stochasticEst}. Specifically, noting that
 $E\left[\vect{n}^Th_{\lambda,K}(\vect{n})\right] =
\trace\left(h_{\lambda,K}(\cdot) \right)$ where $\vect{n}$ is a
random vector $\vect{n} \sim N(0,\matr{I})$, we can approximate
the trace by one such instance of the random vector.
More than one instance of the random variable $\vect{n}$ can be
used for better estimation of the trace by averaging, but in this
work we use just one such realization of $\vect{n}$.
Replacing the
trace in (\ref{eq:GCV_fun}) yields
\begin{eqnarray}
\label{eq:GCV_fun_prac}
\GCV({\lambda,K}) = \frac{\frac{1}{n_\vect{y}}
\norm{\vect{r}_{h_{\lambda,K}(\vect{y})}}_2^2}{\left[1 -
\frac{1}{n_\vect{y}}\vect{n}^T\matr{H}h_{\lambda,K}(\vect{n})\right]^2} & .
\end{eqnarray}
Some extensions of the GCV to non-linear cases were  treated in
\cite{Fu2005Nonlinear,Haber2000GCV,OSullivan1985CrossVal,Vogel2002ComputMethods},
but as far as we know the literature does not offer extensions of
the GCV to non-linear deconvolution algorithms like the one we
address in this work. We suggest using (\ref{eq:GCV_fun_prac}) as
the GCV objective to be minimized for the parameters selection,
despite the non-linearity of $h_{\lambda,K}(\cdot)$.

\begin{figure}[htbp]
\begin{center}
\includegraphics[width=.5\textwidth]{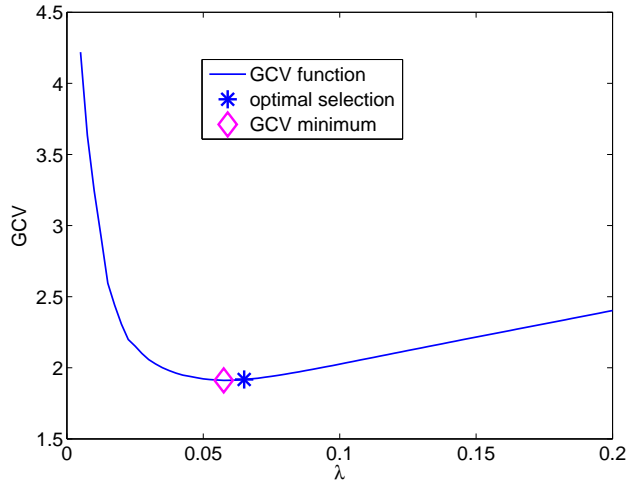}
\end{center}
\caption{The GCV curve as a function of $\lambda$ for the PCD
algorithm, with $ker1$ as blur and with noise power of $\sigma^2=2$
where $K=43$.} \label{fig:GCV_exp1}
\end{figure}

Figure \ref{fig:GCV_exp1} presents an experiment using GCV. For
fixed number of iterations, $K=43$, the GCV as a function of
$\lambda$ is presented in this figure. The value of $\lambda$ is
chosen to be the minimizer of the GCV function. This minimal point
can be found by a line search, but such a procedure may get trapped
in a local minimum of the GCV curve, since the obtained curve is not
necessarily unimodal. Here and in later figures,
the value chosen based on the true MSE is marked by an
asterisk. In this case, as can be seen, the GCV performs well.

\subsection{The L-curve method}
For setting a parameter with the L-curve method, a log of the
squared norm of the result of the algorithm versus the squared
norm of its  residual is plotted over a range of parameter values.
The result is typically a curve with an L shape. The L-curve
criterion is to choose the parameter value that corresponds to the
``corner" of the curve -- the point with maximum curvature
\cite{Hansen93L-curve}. For a parameter of interest $\lambda$, the
curve is defined by the x-axis $X(\lambda) = \log
\norm{r_{h_{\lambda}(\vect{y})}}_2^2$ and the y-axis $Y(\lambda) =
\log \norm{h_{\lambda}(\vect{y})}_2^2$. The curvature is then
given by
\begin{eqnarray}
\label{eq:L-curvature}
\kappa(\lambda) = \frac{X''(\lambda)Y'(\lambda)-X'(\lambda)Y''(\lambda)}
{\left(X'(\lambda)^2 + Y'(\lambda)^2\right)^{3/2}}.
\end{eqnarray}
The value of $\lambda$ is selected to maximize
(\ref{eq:L-curvature}). Since the L-curve can only be sampled at
specific discrete points, the calculation of the second order
derivatives is very sensitive and thus should be avoided.
Following Hansen and O'Leary \cite{Hansen93L-curve}, we apply a local smoothing on
the L-curve points and use the new smoothed points as control
points for a cubic spline. We then calculate the derivatives of
the spline at those points and get the curvature on each of them.
The smoothing is done by fitting five points, where the smoothed
point is the center of them, to a straight line in the least
squares sense.

As in the case of the GCV, according to our knowledge, the L-Curve
has not been applied to non-linear deconvolution methods.
Limitations of this approach for linear deconvolution are
presented in \cite{Vogel2002ComputMethods,Hanke96Limitations}. In
addition, the curvature is very sensitive and thus we can very
easily be diverted from the appropriate value. In addition to this
shortcoming, each calculation of a curvature of a point needs the
values of the L-curve of its neighbors. Therefore, it is hard to
tune the parameters in an efficient way.

\begin{figure}[htbp]
     \centering
     \subfloat[L-curve]{
          \includegraphics[width=.5\textwidth]{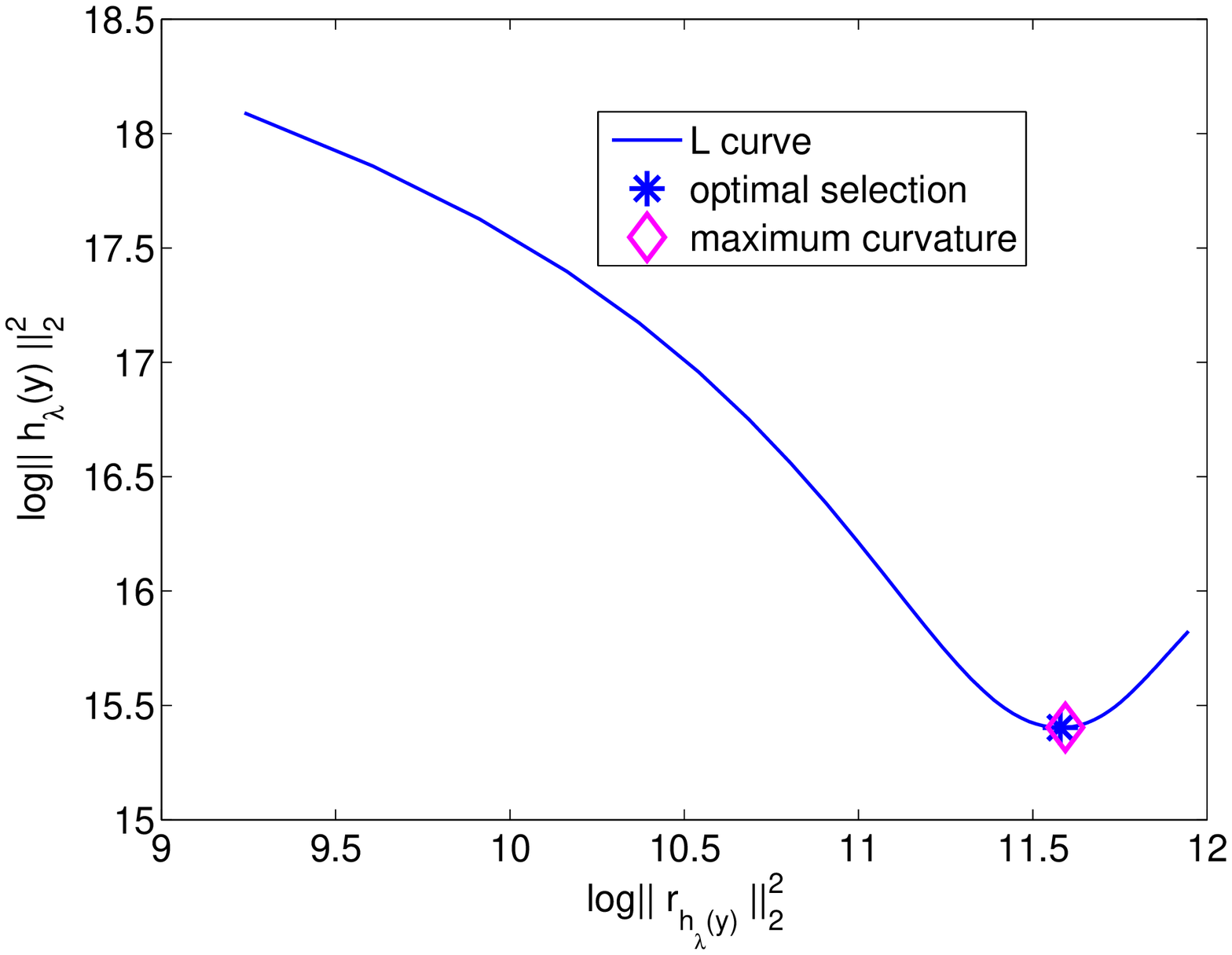}}
     \subfloat[curvature of the L-curve]{
          \includegraphics[width=.5\textwidth]{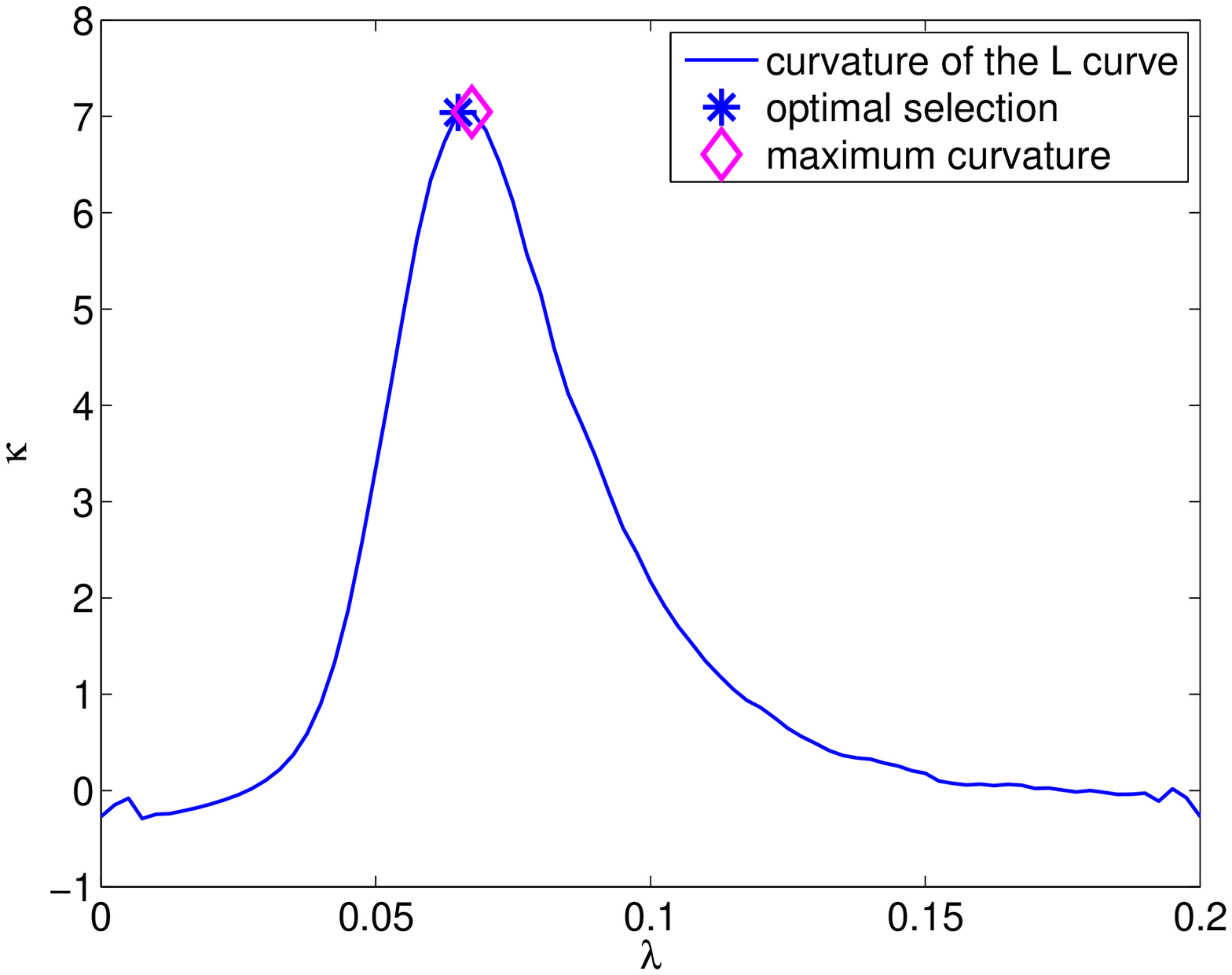}}
     \caption{The L-curve results for setting $\lambda$ for the PCD algorithm
     with $ker1$, $\sigma^2=2$, and fixing $K=43$.
     The L-curve is on the top and its curvature is on bottom.}
     \label{fig:L-curve_exp1}
\end{figure}

Figure \ref{fig:L-curve_exp1} presents the experiment results using
the L-curve method. This figure shows both the L-curve (top) and its
curvature (bottom) for a fixed $K=43$. The location of the chosen
parameter $\lambda$ is marked by a diamond. As can be seen, the
$\lambda$ value chosen as the curvature maximizer, in this case, is very close to
the ideal value.

\subsection{The Discrepancy Principle}

A very intuitive parameter tuning idea is to use the discrepancy
principle.  Based on the noise properties, the original image
approximately satisfies $\|\matr{H}\vect{\vect{x}}- \vect{y}
\|_2^2 \approx n_\vect{y} \sigma^2$. The discrepancy principle
assumes that the same should hold for the reconstructed image and
thus selects the parameters based on the assumption that
\begin{eqnarray}
\label{eq:disc_princ}
\norm{\matr{H}h_{\lambda,K}(\vect{y})- \vect{y} }_2^2 \approx n_\vect{y} \sigma^2.
\end{eqnarray}
The iteration number, $K$, is chosen to be the first iteration in
which $\norm{\matr{H}h_K(\vect{y})- \vect{y} }_2^2 < n_\vect{y}
\sigma^2$.

\begin{figure}[htbp]
\begin{center}
\includegraphics[width=.5\textwidth]{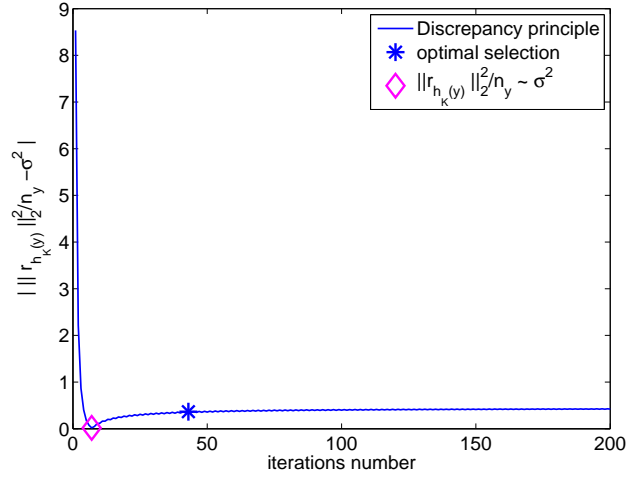}
\end{center}
\caption{The discrepancy principle for selecting the number of
iterations for the PCD algorithm with $ker1$, $\sigma^2=2$, and
fixing $\lambda=0.065$. } \label{fig:discrepancy_exp1}
\end{figure}

Experimenting with the discrepancy principle,
Fig.~\ref{fig:discrepancy_exp1} presents the value of
$\abs{\|\matr{H}h_K(\vect{y})- \vect{y} \|_2^2 -n_\vect{y}
\sigma^2}$ as a function of $K$, where $\lambda = 0.065$ is fixed.
The discrepancy principle selects the parameter value that
achieves the minimal value on the graphs. As can be seen, although
this method is very appealing, in many instances like this one, it
tends to select a value of $K$ far from the optimal one.

In this section we have described three parameter tuning methods: The GCV,
the
L-curve and the discrepancy principle. The discrepancy principle, which is
commonly used as a stopping criterion, tends to fail in many cases, as can be
seen in this section.
As for the GCV and the L-curve methods, as far as
we know, they were not applied before to non-linear deconvolution algorithms
of
the kind we use here. While their deployment is reasonable, their
performance tends to be overly
sensitive, possibly leading to poor results, as will be shown in the results
section. The fact that these three methods may give
good parameter tuning in part of the cases may be
explained by the fact that these methods do not aim at minimizing
the MSE but use different considerations for the tuning. This
leads us naturally to consider ways to select the parameters
directly based on an estimated value of the MSE.


\section{GSURE and Projected GSURE}
\label{sect:sure_gsure}

In this section we discuss an unbiased estimator for the MSE
obtained by any reconstruction algorithm. This MSE estimate will
later be used for parameter selection. The Stein unbiased risk
estimator (SURE) was proposed in
\cite{Stein73Estimate--mean,Stein81Estimate--mean} as an estimate
of the MSE in the Gaussian denoising problem, namely, when
$\matr{H}={\bf I}$ and the noise is i.i.d. and Gaussian. This idea
was then extended in \cite{Eldar09GSURE} to handle more general
inverse problems of the form discussed here, and a wider class of
noise distributions. We refer to the later MSE estimate as the
generalized SURE (GSURE).

The essential concept behind the GSURE principle is to seek an
unbiased estimate for the MSE that is only a function of the
estimate $h(\vect{y})$ and $\vect{y}$ (up to a constant which may
depend on $\vect{x}$ but is independent of $h(\vect{y})$). Denote
such an MSE estimate by $\eta(\vect{\hat{ \vect{x}}},\vect{y})$.
Then,
\begin{equation}\label{eq:unbiased}
E\left[\eta(\vect{\hat x},\vect{y})\right] = E\left[ \|\vect{\hat
x}-\vect{x}\|_2^2 \right] = E\left[ \|h(\vect{y})-\vect{x}\|_2^2
\right].
\end{equation}
If we have access to such a function $\eta(\vect{\hat
x},\vect{y})$ that estimates the MSE, while being also dependent
on a set of parameters $\Theta$ (that control the reconstruction
 performance), then we may choose the values of $\Theta$ so as to
minimize this function, and in that way aim at minimizing the true
MSE as a consequence.

For the case where $\matr{H}$ has full column rank, the GSURE is
given by Eldar \cite{Eldar09GSURE}
\begin{equation} \label{eq:GSURE}
\eta(h(\vect{u}),\vect{y}) =c_1 + \norm{h(\vect{u})}^2_2 -
2\vect{x}_{ML}^Th(\vect{u})+ 2\nabla_\vect{u} \cdot h(\vect{u}).
\end{equation}
Here, $\vect{u}=(1/\sigma^2)\matr{H}^T\vect{y}$ is the sufficient
statistic  for the model (\ref{eq:blur}),  $\nabla$ is the
divergence operator
\begin{equation}
\nabla_{\vect{z}}h(\vect{u}) =
\sum_{i=1}^n\frac{dh(\vect{u})[i]}{d\vect{z}[i]}
=\trace\left(\frac{dh(\vect{u})}{d\vect{z}}\right),
\end{equation}
and $c_1$ is a constant independent of $h(\vect{u})$. Given that
our goal is to minimize the MSE by setting $h(\vect{u})$, the
constant $c_1$ is irrelevant for the minimization process. The
term $\vect{x}_{ML}$ stands for the Maximum Likelihood (ML)
estimator
--- the minimizer of
\begin{equation}
\label{eq:LS_obj} \min_\vect{z}\norm{\vect{y}-\matr{H}\vect{z}}.
\end{equation}
If $\matr{H}^T\matr{H}$ is invertible, then this estimate is given
by
\begin{equation} \label{eq:ML_est}
\vect{x}_{ML}= (\matr{H}^T\matr{H})^{-1}\matr{H}^T \vect{y}.
\end{equation}

The MSE estimate presented in (\ref{eq:GSURE}) is valid only if
$\matr{H}^T\matr{H}$ is invertible. When this condition is not
satisfied, $\matr{H}$ has a nontrivial null space. Therefore, we
have no information about the components of $\vect{x}$ that lie in
the null space $\mathcal{N}(\matr{H})$. Thus we can estimate only
the projected MSE on the range of $\matr{H}^T$ (denoted by
$\mathcal{R}(\matr{H}^T)$), which is equal to the orthogonal
complement of the null space $\mathcal{N}(\matr{H})^\perp$.
Therefore, instead of attempting to minimize an unbiased estimate
of the MSE, we consider an unbiased estimate of the projected MSE:
$E\left[
\matr{P}_{\mathcal{R}(\matr{H}^T)}\|h(\vect{u})-\vect{x}\|_2^2
\right]$, where
\begin{eqnarray}
\label{eq:proj_range}
  \matr{P}_{\mathcal{R}(\matr{H}^T)} &=& \matr{H}^T(\matr{HH}^T)^\dag
  \matr{H},
\end{eqnarray}
is the orthogonal projection onto
$\mathcal{R}(\matr{H}^T)=\mathcal{N}(\matr{H})^\perp$. Here
 $\dag$
denotes the Moore-Pernose pseudo inverse.

As shown in \cite{Eldar09GSURE}, an unbiased estimator for the
projected MSE is given by the same formula as (\ref{eq:GSURE}),
but applied on the projected estimate $h(\vect{u})$:
\begin{eqnarray}
\label{eq:PGSURE}
\nonumber
\eta_{\matr{P}_{\mathcal{R}(\matr{H}^T)}}(h(u),y) & = & c_2 +
\norm{\matr{P}_{\mathcal{R}(\matr{H}^T)}h(\vect{u})}_2^2 -
2\vect{x}_{ML}^T  h(\vect{u})\\&& + 2\nabla_\vect{u} \cdot
(\matr{P}_{\mathcal{R}(\matr{H}^T)}h(\vect{u})).
\end{eqnarray}
Again, $c_2$ is a constant independent of $h(\vect{y})$. Here the
ML estimate is chosen as the minimum-norm minimizer of
(\ref{eq:LS_obj}), given by
\begin{equation}
\label{eq:ML_est_rank_def}
\vect{x}_{ML}  = \matr{H}^T(\matr{HH}^T)^\dag \vect{y}.
\end{equation}


\section{Parameter Tuning with GSURE}
\label{sect:param_set_gsure}

The generalization of SURE provides the possibility of using this
estimator for the case of general $\matr{H}$, and specifically,
for the deblurring and image scale-up problems. The tuning will be
demonstrated for the SSF and PCD algorithms. As seen in Section
\ref{sect:iterative_shrinkage}, these methods have mainly two
unknown parameters, the iterations number $K$ and the thresholding
value $\lambda$. Vonesch, Ramani and Unser, in
\cite{Vonesch08GSURE}, have already addressed this problem for the
case where the operator $\matr{H}$ has full column rank. In their method,
referred to as the {\em global method}, one parameter is being set
at a time, given the other. We extend this for general $\matr{H}$
using the projected GSURE, showing the advantage of its usage over
the regular GSURE, and over the conventional parameter setting
methods that were presented in Section
\ref{sect:ConvParametrTuningMethods}.

Though the global method for parameter setting provides us with a
good estimation of the parameters as demonstrated in the results
section, it sets only one parameter given the other. We present an
additional approach that sets both together. The parameter
$\lambda$ takes on a different value in each iteration of the
algorithm in a greedy fashion. This way, we choose $K$ together
with $\lambda$, setting it to be the point where the estimated MSE
stops decreasing. This method has the same computational cost for
choosing two parameters as the global method has for selecting
only one, without degrading the MSE performance. We also suggest a
{\em greedy look-ahead method} that aims at better reconstruction
results at the expense of higher complexity. We compare between
these algorithms and the global method demonstrating the advantage
of using these approaches for parameter tuning when both $\lambda$
and $K$ are unknown.

\subsection{The Global Method using GSURE}
\label{sect:param_set_global}

In order to use GSURE with the iterated shrinkage algorithms,
described in Section \ref{sect:iterative_shrinkage}, we need to
calculate the ML estimator, the derivative of the reconstruction
algorithm with respect to $\vect{u}$, and the projection operator
in the rank-deficient case. In our experiments, the ML estimator
and the projection matrix are easily calculated by exploiting the
fact that the blur operators considered are block circulant. As
such, they are diagonalized by the 2D discrete Fourier transform
(DFT). This helps both for the deblurring and the scale-up
problems we handle.

To calculate the derivatives with respect to $\vect{u}$, we  first
reformulate the iterative equation as a function of $\vect{u}$
instead of $\vect{y}$. Rewriting (\ref{eq:ssf_iter}) leads to
\begin{equation} \label{eq:ssf_iter_u}
\alphabf_{k+1} =
\mathcal{S}_{\rho,\lambda/c}\left(\frac{1}{c}(\matr{D})^T(\sigma^2\vect{u}-
\matr{H}^T\matr{HD}\alphabf_k)+\alphabf_k\right),
\end{equation}
where $\hat{\vect{x}} = h_{\lambda,K}(\vect{u}) =
\matr{D}\alphabf_{K}$ as in (\ref{eq:estimator}).
The derivatives of $\hat{\vect{x}}$ are
calculated recursively by first noticing that
\begin{eqnarray} \label{eq:ssf_iter_deriv}
\nonumber
\frac{d\alphabf_{k+1}}{d\vect{u}} & = &
 \mathcal{S}_{\rho,\lambda/c}'\left(\frac{1}{c}\matr{D}^T(\sigma^2\vect{u}-\matr{H}^T
 \matr{HD}\alphabf_{k})+
 \alphabf_{k}\right) \cdot \\
&& \left[\frac{\sigma^2}{c}\matr{D}^T -
\frac{1}{c}(\matr{HD})^T\matr{HD}\frac{d\alphabf_{k}}{d\vect{u}} +
\frac{d\alphabf_{k}}{d\vect{u}} \right],
\end{eqnarray}
where $\mathcal{S}_{\rho,\lambda/c}'(\cdot)$ is an element wise
derivative of the shrinkage function, organized as a diagonal
matrix. From here, the divergence of the estimator can be directly
obtained by multiplying the above by $\matr{D}$ from the left,
gathering the diagonal of the matrix and summing it up:
\begin{eqnarray}
\label{eq:divergence}
\nabla_{\vect{u}}h_{\lambda,K}(\vect{u})
= \trace\left(\frac{h_{\lambda,K}(\vect{u})}{d\vect{u}}\right) =
\trace\left(\matr{D}\frac{d\alphabf_{K}}{d\vect{u}}\right).
\end{eqnarray}
It is impracticable to calculate the Jacobian of
(\ref{eq:ssf_iter_deriv}) in high dimensions. Instead we use the
trace estimator that was used for the GCV in Section \ref{sect:GCV}.
Following Vonesch \etal \cite{Vonesch08GSURE}
we choose a random vector $\vect{n} \sim N(0,\matr{I})$. Using
this estimator we iterate over
$\frac{d\alphabf_{k}}{d\vect{u}}\vect{n}$ , which can be
calculated instead of $\frac{d\alphabf_{k}}{d\vect{u}}$, leading
 to the following iterative formula
\begin{eqnarray} \label{eq:ssf_iter_deriv_est}
\nonumber
\frac{d\alphabf_{k+1}}{d\vect{u}}\vect{n} =
 S_{\rho,\lambda/c}'\left(\frac{1}{c}\matr{D}^T(\vect{u}-\matr{H}^T
 \matr{HD}\alphabf_{k})+
 \alphabf_{k}\right) \cdot \\
 \left[\frac{\sigma^2}{c}\matr{D}^T\vect{n} -
\frac{1}{c}(\matr{HD})^T\matr{HD}\frac{d\alphabf_{k}}{d\vect{u}}\vect{n} +
\frac{d\alphabf_{k}}{d\vect{u}}\vect{n} \right].
\end{eqnarray}

For the estimation of the MSE for the PCD algorithm we need to
calculate the trace as was done for the SSF. We rewrite the PCD
iteration in terms of $\vect{u}$
\begin{eqnarray}
\label{eq:PCD_iter_u} \alphabf_{k+1}  & = & \alphabf_k +
\mu(\vect{v}_k - \alphabf_k) \nonumber \\ \nonumber & = &
(1-\mu)\alphabf_k
 \\ && + \mu\left(\mathcal{S}_{\rho,\matr{W}\lambda}
\left( \matr{W} \matr{D}^T(\sigma^2\vect{u} -
\matr{H}^T\matr{HD}\alphabf_k ) + \alphabf_k \right)
\right).
\end{eqnarray}
The derivative with respect to $\vect{u}$ is given by
\begin{eqnarray}
\label{eq:PCD_iter_deriv} \nonumber \frac{d\alphabf_{k+1}}{du} & = &
(1-\mu)\frac{d\alphabf_{k}}{du}
\\ \nonumber
&& + \mu \mathcal{S}'_{\rho,\matr{W}\lambda} \left( \matr{W}\matr{D}^T
(\sigma^2\vect{u} - \matr{H}^T\matr{HD}\alphabf_k ) + \alphabf_k \right)
  \cdot \\ && \left[\matr{W}\matr{D}^T(\sigma^2\matr{I}
- \matr{H}^T\matr{HD}\frac{d\alphabf_{k}}{d\vect{u}} )  + \frac{d\alphabf_{k}}{d\vect{u}} \right].
\end{eqnarray}
Using randomization we get the following iterative formula for
estimating the trace value,
\begin{eqnarray}
\label{eq:PCD_iter_deriv_est}
\nonumber
\frac{d\alphabf_{k+1}}{du} \vect{n} & = &
(1-\mu)\frac{d\alphabf_{k}}{du}\vect{n} \\
&& \nonumber + \mu \mathcal{S}_{\rho,\matr{W}\lambda}'\left( \matr{W}\matr{D}^T
(\sigma^2\vect{u} - \matr{H}^T\matr{HD}\alphabf_k )  + \alphabf_k \right) \cdot \\ && \left[\matr{W}\matr{D}^T(\sigma^2\vect{n}
 - \matr{H}^T\matr{HD}\frac{d\alphabf_{k}}{d\vect{u}}\vect{n} )   + \frac{d\alphabf_{k}}{d\vect{u}}\vect{n}    \right]
.
\end{eqnarray}

Now that we hold a complete expression for the GSURE MSE estimate,
$K$ and $\lambda$ can be chosen by minimizing it. This strategy
was developed in \cite{Vonesch08GSURE} and is referred to as the
global method. In the full-rank case, the projection matrix is the
identity and the projected GSURE coincides with the regular GSURE.
Thus we can use the projected GSURE equation in (\ref{eq:PGSURE})
for both cases. For a fixed and pre-chosen number of iterations
$K$, the $\lambda$ which minimizes the GSURE expression is chosen.
Repeating this process for various values of $K$, one can minimize
the overall global MSE with respect to both $K$ and $\lambda$. As
an analytical minimization of the GSURE is hard to achieve, we use
a golden section search \cite{Kiefer53Sequential--minimax}. We
initialize these algorithms by
\begin{eqnarray}
\label{eq:init_alpha} \alphabf_0 = (\matr{HD})^T\vect{y},
 \end{eqnarray}
and
\begin{eqnarray}
\label{eq:init_alpha_deriv} \frac{d\alphabf_{0}}{d\vect{u}} =
\sigma^2\matr{D}^T.
\end{eqnarray}

Denoting by $T$ the GSURE calculation time (per-iteration),
$n_{gs}$ the golden-section number of iterations, and $n_{is}$ the
number of iterations of the iterated shrinkage algorithm, the time
complexity of the global method for setting  $\lambda$, when $K$
is fixed, is $O(n_{is}n_{gs}T)$. For setting $K$ another golden
section can be performed over the number of iterations.


\subsection{The Greedy Method using GSURE}
\label{sect:param_set_greedy}

In the greedy method, instead of choosing one  global $\lambda$
for all the iterations, we take a different route. Turning to a
local alternative, the value of $\lambda$ can be determined as the
minimizer of the estimated MSE of the current iteration. In this
strategy, $K$ is set together with $\lambda$ as the point where
the estimated MSE (almost) stops decreasing. This allows us to
avoid setting each of the parameters separately, as done in the
global method. The algorithm proposed is the following:
\begin{itemize}
\item Initialize $\alphabf_0$ and calculate its derivative w.r.t.
$\vect{u}$.
\item Repeat:
\begin{enumerate}
  \item Set $\lambda_k^* =
  \arg\min_\lambda(\eta_{\matr{P}_{\mathcal{R}(\matr{H}^T)}} (h_\lambda(\vect{u},\alphabf_{k-1})))$.
  \item Perform the iterations in (\ref{eq:PCD_iter_u}) and (\ref{eq:PCD_iter_deriv_est}) (PCD case) or (\ref{eq:ssf_iter_u}) and (\ref{eq:ssf_iter_deriv_est}) (SSF case) for the calculation of $\alphabf_{k}$ and $\frac{d\alphabf_{k}}{d\vect{u}}\vect{n}$ using $\lambda_k^*$.
  \item Compute \\ $\MSE_k^* = \eta_{\matr{P}_{\matr{P}_{\mathcal{R}(\matr{H}^T)}}}(\matr{D}\alphabf_{k}, \vect{n}^T\matr{D}\frac{d\alphabf_{k}}{d\vect{u}}\vect{n},\vect{x}_{ML})$ using (\ref{eq:PGSURE}), (\ref{eq:estimator}), (\ref{eq:divergence}), (\ref{eq:proj_range}) and (\ref{eq:ML_est}) or (\ref{eq:ML_est_rank_def}).
  \item If $\MSE_{k-1}^*-\MSE_{k}^*\le \delta$, stop.
\end{enumerate}
    \item $\hat{\vect{x}} = \matr{D}\alphabf_k$.
\end{itemize}

The complexity of the greedy algorithm is also $O(n_{is}n_{gs}T)$.
When one of the parameters is fixed in the global method and we
need to set only the other parameter we get the same complexity in
the two approaches. In the case where both parameters are to be
set, the greedy technique has an advantage since it chooses the
number of iterations along with the parameter $\lambda$. In
contrast, in the global method the number of iterations is either
pre-chosen and thus suboptimal, or searched using yet another
golden-search for optimizing $K$ increasing the overall
complexity.

To further decrease the MSE,  we can modify the greedy algorithm
by introducing a look-ahead option. One of the problems of the
greedy strategy is that it minimizes the MSE of the current
iteration but can harm the overall MSE. Thus, instead of choosing
$\lambda$ that minimizes the estimated MSE of the current
iteration, we select it to minimize the estimated MSE of $r$
iterations ahead, assuming that these $r$ iterations are performed
using the greedy approach described above. This change provides a
look-ahead of $r$ iterations, formulated as the following
algorithm:
\begin{itemize}
\item Initialize $\alphabf_0$ and calculate its derivative w.r.t. $\vect{u}$.
\item Repeat:
\begin{enumerate}
  \item Set $\lambda_k^*$ by minimizing the estimated MSE $r$ iterations ahead
  using the above-described greedy algorithm.
  \item Perform a single iteration as in (\ref{eq:PCD_iter_u}) and (\ref{eq:PCD_iter_deriv_est}) (PCD case) or (\ref{eq:ssf_iter_u}) and (\ref{eq:ssf_iter_deriv_est}) (SSF case) for the calculation of $\alphabf_{k}$ and $\frac{d\alphabf_{k}}{d\vect{u}}\vect{n}$ using $\lambda_k^*$.
  \item Compute \\ $\MSE_k^* = \eta_{\matr{P}_{\matr{P}_{\mathcal{R}(\matr{H}^T)}}}(\matr{D}\alphabf_{K}, \vect{n}^T\matr{D}\frac{d\alphabf_{K}}{d\vect{u}}\vect{n},\vect{x}_{ML})$ using (\ref{eq:PGSURE}), (\ref{eq:estimator}), (\ref{eq:divergence}), (\ref{eq:proj_range}) and (\ref{eq:ML_est}) or (\ref{eq:ML_est_rank_def}).
  \item If $\MSE_{k-1}^*-\MSE_{k}^*\le \delta$, stop.
\end{enumerate}
    \item $\hat{\vect{x}} = \matr{D}\alphabf_k$.
\end{itemize}
In step 1, for each test $\lambda$ in the golden-section, $r$
iterations of the greedy method are performed, as described for
the greedy algorithm.  Finally, $\lambda$ of the current iteration
is chosen such that the estimated MSE of the last $r$th greedy
iteration is minimized. The time complexity of the $r$ look-ahead
greedy algorithm is $n_{is}(n_{gs})^2rT$, which is $n_{gs}r$ times
slower than the other two methods. However, this is partially
compensated for in some cases by getting smaller $n_{is}$ due to
faster convergence. Furthermore, the resulting MSE is often lower,
as we illustrate in the results section.


\section{Results}
\label{sect:results}

The greedy methods were previously presented, with preliminary results on low dimension signals, in \cite{Giryes08Automatic--parameter}.
In this work we extended their use for images. In this section we present a set of experiments that demonstrate
parameter tuning for the deblurring and the scale-up problems,
using the PCD and the SSF algorithms. The representative
experiments chosen correspond to different scenarios, which enable
a demonstration of the various algorithms and their behavior. We
should note that our work included various more tests, which are
not brought here because of space limitation, and which lead to
the same final conclusions. Table \ref{table:experiments} lists
the five tests performed. The tests we present divide into two
groups: (i) preliminary ones that demonstrate the superiority of
the projected GSURE over all the other alternatives, and then (ii)
advanced tests that compare the global and the greedy methods.

\begin{table*}[htbp]
\centering
\begin{tabular}{||c|c|c|c|c|c||}
\hline \hline
Problem & Blur & $\sigma^2$ & Decimation & Algorithm & Cond-\# \\
\hline  \hline
1. Deblurring & $ker1: (1+x_1^2+x_2^2)^{-1}$ & $2$ or $8$ & none & PCD & $77$ \\
~ & $-7 \le x_1,x_2\le 7$ & ~ & ~ & ~ & ~ \\
\hline
2. Deblurring & $ker2: 1/81$ & $0.308$ & none & PCD & $2.5e+5$ \\
~ & $-4 \le x_1,x_2\le 4$ & ~ & ~ & ~ & ~ \\
\hline
3. Deblurring & $ker3: [1 , 4 , 6 , 4 , 1]$ & $49$ & none & PCD & $\infty$ \\
~ & separable & ~ & ~ & ~ & ~ \\
\hline
4. Scale-Up & $ker4: [1, 2, 1]/4$ & $49$ & $2:1$ & SSF & $\infty$ \\
~ & separable & ~ & in each axis & ~ & ~ \\
\hline
5. Scale-Up & $ker5: [1, 1]/2$ & $16$ & $2:1$ & SSF & $\infty$ \\
~ & separable & ~ & in each axis & ~ & ~ \\
\hline \hline
\end{tabular} \caption{A list of the experiments
presented.} \label{table:experiments}
\end{table*}

\subsection{Preliminary Tests}

\begin{figure}[htbp]
     \centering
          \includegraphics[width=.5\textwidth]{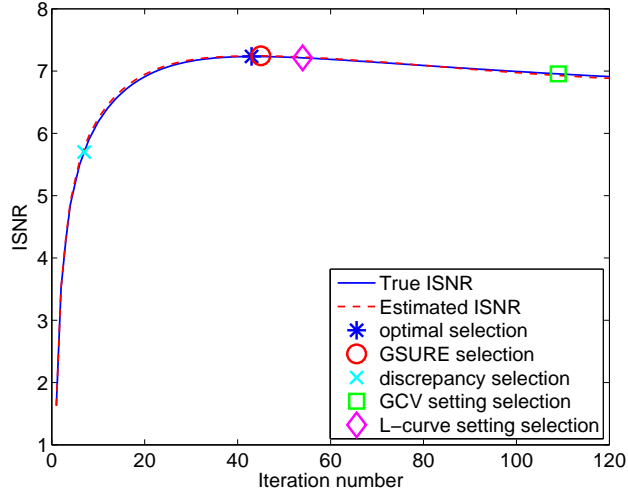}
     \caption{The GSURE-estimated and the true ISNR as  a function of the
iterations number for Problem-1 with $\sigma^2=2$ and fixing
$\lambda =0.065$. The selection of the iterations number based on
the various discussed methods is marked.}
     \label{fig:iter_GSUREest_exp1}
\end{figure}

\begin{figure}[htbp]
     \centering
          \includegraphics[width=.5\textwidth]{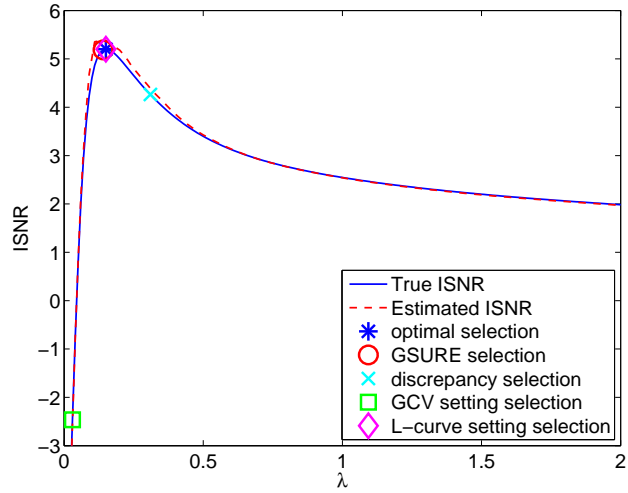}\\
     \caption{The GSURE-estimated and the true ISNR as a function of
     $\lambda$ for Problem-1 with $\sigma^2=8$ and fixing $K=27$.
     The selection of $\lambda$ based on the various discussed methods is marked.}
     \label{fig:lambda_GSUREest_exp18}
\end{figure}

We start by demonstrating the core ability of the GSURE and the
classical techniques to choose the required parameters. In Fig.~
\ref{fig:iter_GSUREest_exp1} we present the GSURE ISNR
estimation\footnote{The reference MSE is obtained by considering
$\vect{y}$ as the estimate.} for Problem-1 applied on {\tt
Cameraman} with $\sigma^2=2$ as a function of the iterations.
Similarly, Fig.~\ref{fig:lambda_GSUREest_exp18} presents the GSURE
ISNR as a function of $\lambda$ for the same problem with
$\sigma^2=8$. In both cases, the other parameter is kept fixed,
choosing it to be the optimal one based on the true MSE. It can be
observed in both graphs that the GSURE gives a very good
estimation for the MSE and thus also a high-quality estimate of
both $\lambda$ and the number of iterations $K$. It can also be
seen that it selects the parameters to be much closer to the
optimal selection, compared to all the conventional methods. Among
these conventional methods, the L-curve gives the best
approximation for the parameters. However, it is hard to compute
in an efficient way. The other methods approximate $\lambda$
relatively well but fail in choosing $K$.

Repeating the experiment reported in
Fig.~\ref{fig:iter_GSUREest_exp1} on the image {\tt Boat} and
fixing $\lambda = 0.0725$, Table \ref{tbl:boat_iter_exp1}
summarizes the ISNR reconstruction results for setting $K$. It can
be seen again that the GSURE gives the best reconstruction
results, being approximately the same as those achieved by the
optimal selection.

\begin{table}[htbp]
  \centering
\begin{tabular}{||c|c||}
  \hline \hline
  Parameter Tuning Method & ISNR \\
  \hline
  Optimal Selection & 6.461dB \\
  GSURE & 6.458dB \\
  L-curve & 6.43dB \\
  GCV & 5.21dB \\
  Discrepancy Principle & 5.47dB \\
  \hline \hline
\end{tabular}
\caption{ISNR deblurring results for Problem-1 on {\tt Boat}  with
$\sigma^2=2$, fixing $\lambda = 0.0725$ and setting $K$ using
various methods.} \label{tbl:boat_iter_exp1}
\end{table}

\begin{figure}[htbp]
     \centering
          \includegraphics[width=.5\textwidth]{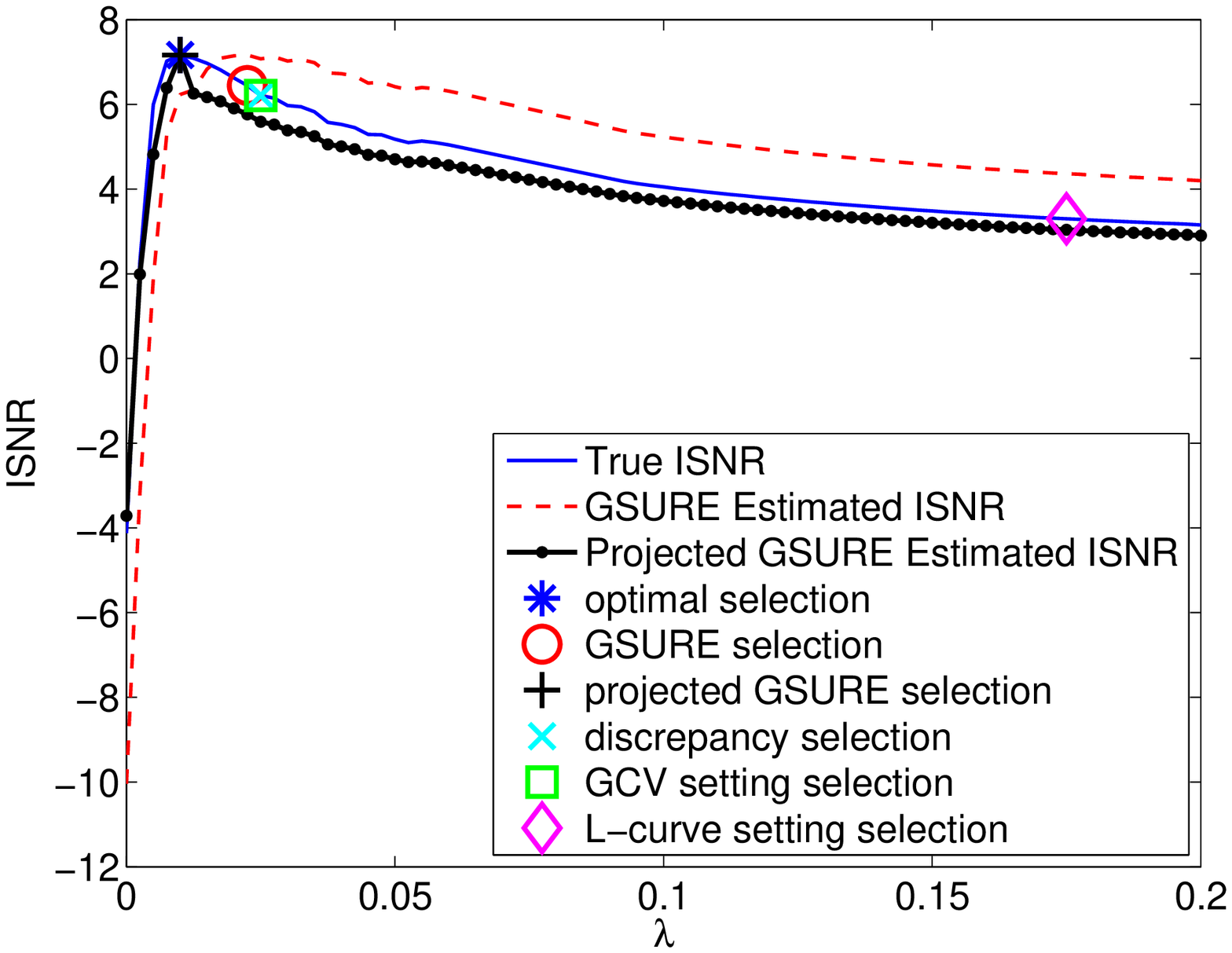}
     \caption{The true and estimated ISNR as a function of
     $\lambda$, using the GSURE and its projected version. Applied on Problem-2,
     the iterations number is set to
     $K = 150$, and $\lambda$ is estimated using the various
     proposed methods.}
     \label{fig:lambda_GSUREest_exp2}
\end{figure}

\begin{figure}[htbp]
     \centering
          \includegraphics[width=.5\textwidth]{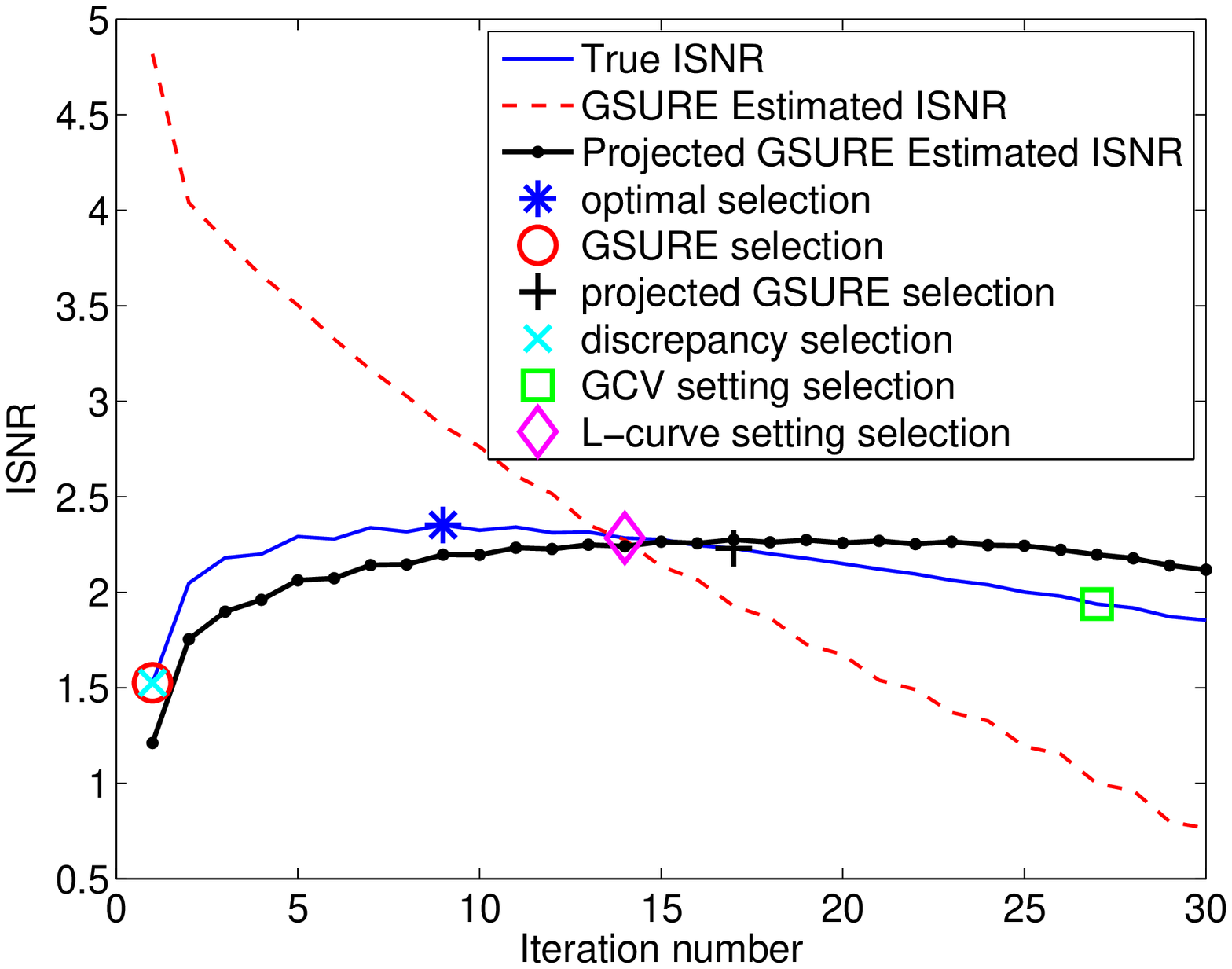}
     \caption{The true and estimated ISNR as a function of
     $K$, using the GSURE and its projected version. Applied on Problem-3,
     $\lambda$ is set to $\lambda = 0.67$, and $K$ is estimated using the various
     proposed methods.}
     \label{fig:iter_GSUREest_exp3}
\end{figure}

In Figs.~\ref{fig:lambda_GSUREest_exp2} and
\ref{fig:iter_GSUREest_exp3} a comparison is made between the ISNR
estimation using the regular and the projected GSURE. The problem
considered in the first graph is Problem-2 and in the second is Problem-3,
where both are applied on {\tt Cameraman}. The selected
parameters of the various methods are marked. The first graph is a
function of $\lambda$ and the second is a function of the
iterations $K$. It can be seen that for both operators the
estimation of $\lambda$ and the iterations number using the
projected GSURE is reasonable and always better than using the
regular GSURE. Also, we see that in the case where we have a blur
with real null space we get a totally unreliable estimation of the
ISNR with the regular GSURE. We also see that L-curve fails this
time in estimating $\lambda$.

\begin{figure}
     \centering
     \subfloat[original image.]{
          \includegraphics[width=.24\textwidth]{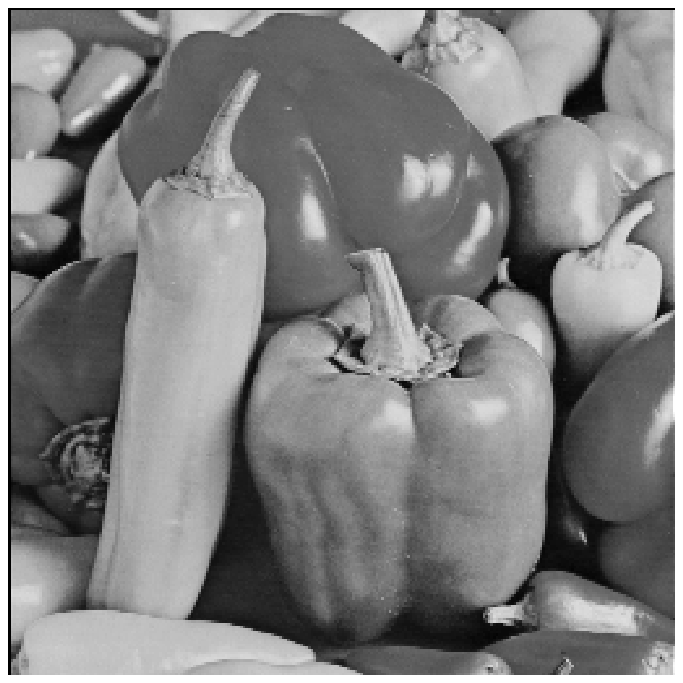}}
     \hspace{.0in}
     \subfloat[blurred image.]{
          \includegraphics[width=.24\textwidth]{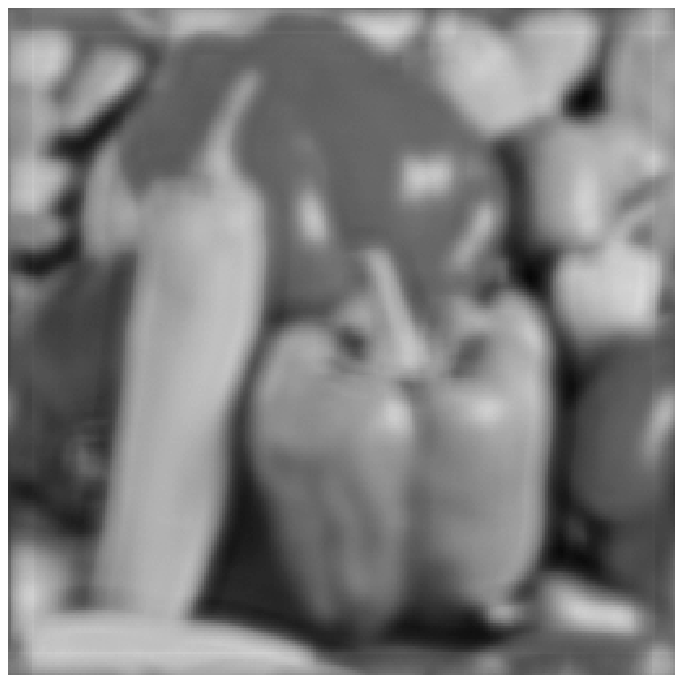}}
     \vspace{.0in}
     \subfloat[optimal selection. ISNR = 8.82 dB.]{
          \includegraphics[width=.24\textwidth]{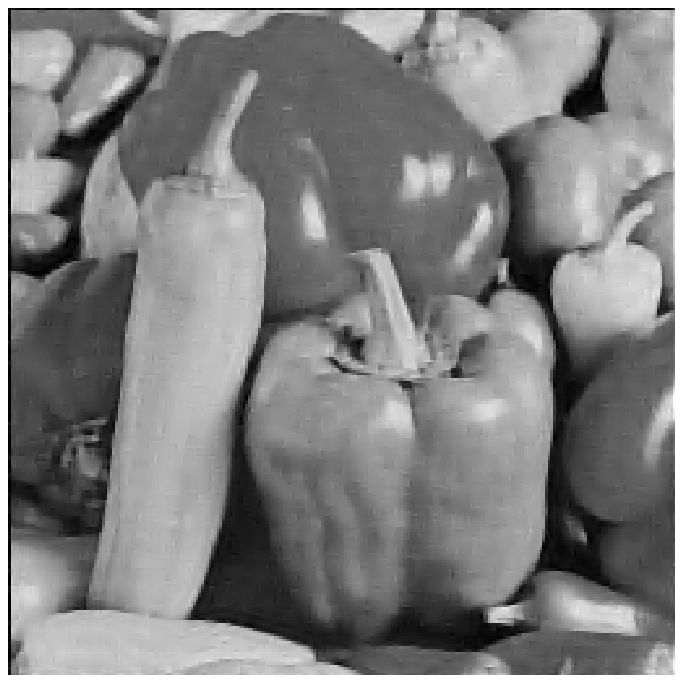}}\\
     \hspace{.0in}
     \subfloat[GCV selection. ISNR = 8.05 dB.]{
          \includegraphics[width=.24\textwidth]{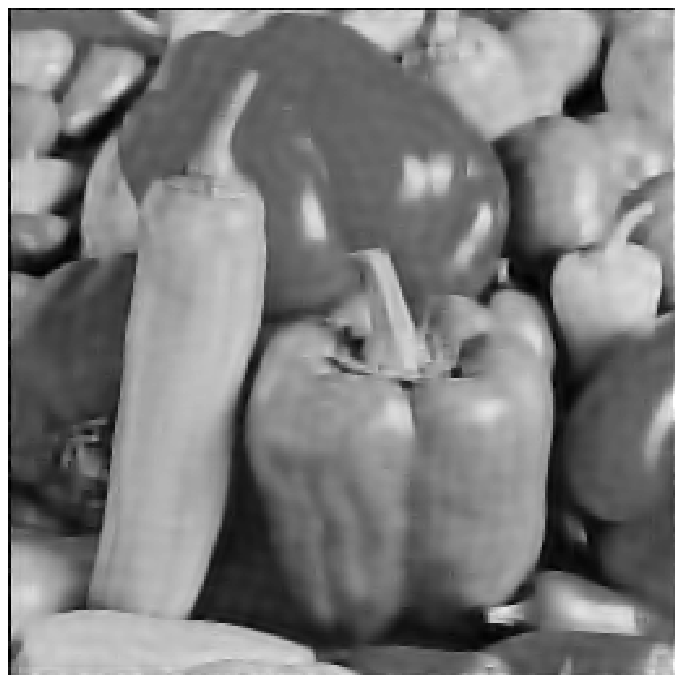}}
     \vspace{.0in}
     \subfloat[L-curve selection. ISNR = 7.64 dB.]{
           \includegraphics[width=.24\textwidth]{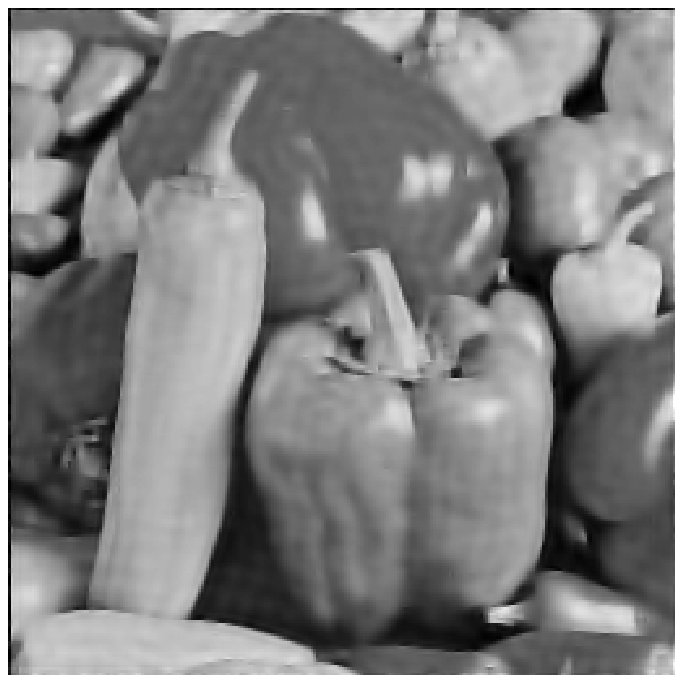}}
     \hspace{.0in}
     \subfloat[discrepancy principle selection. ISNR = 8.21 dB.]{
          \includegraphics[width=.24\textwidth]{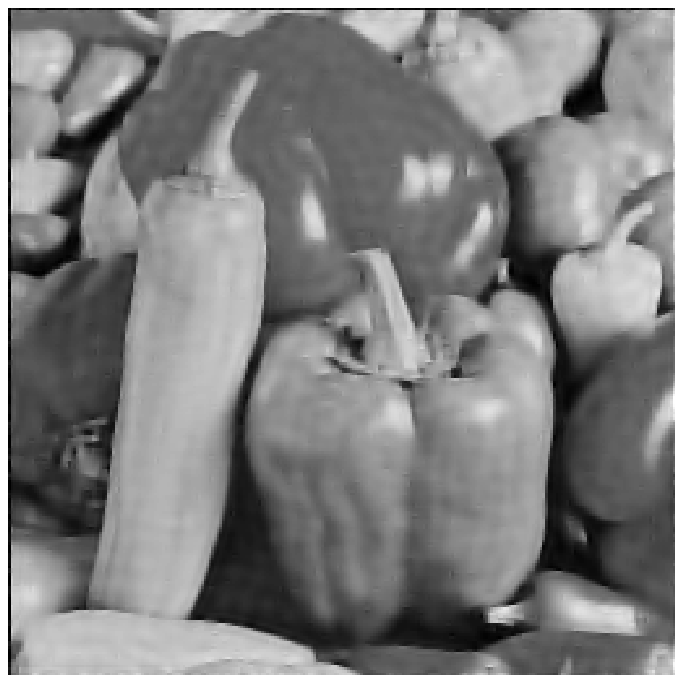}}\\
     \vspace{.0in}
     \subfloat[GSURE selection. ISNR = 8.05 dB.]{
          \includegraphics[width=.24\textwidth]{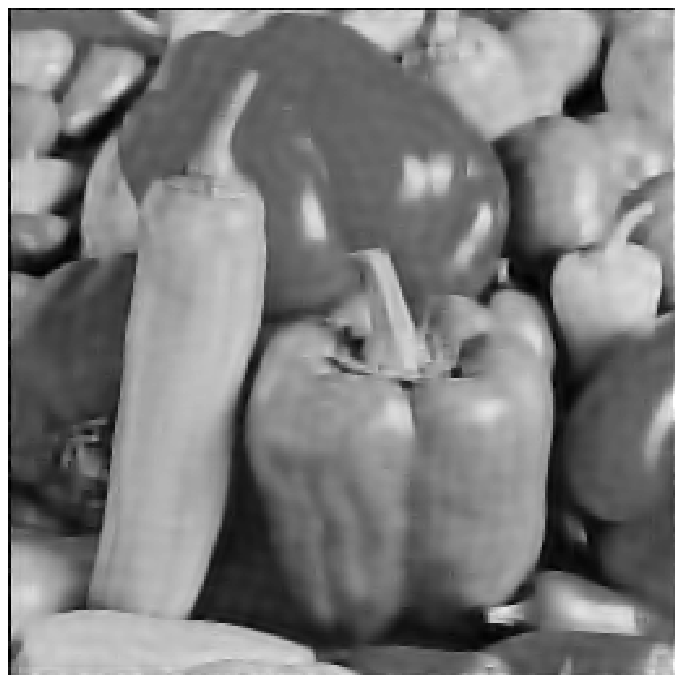}}
     \hspace{.0in}
     \subfloat[projected GSURE selection. ISNR = 8.82 dB ]{
          \includegraphics[width=.24\textwidth]{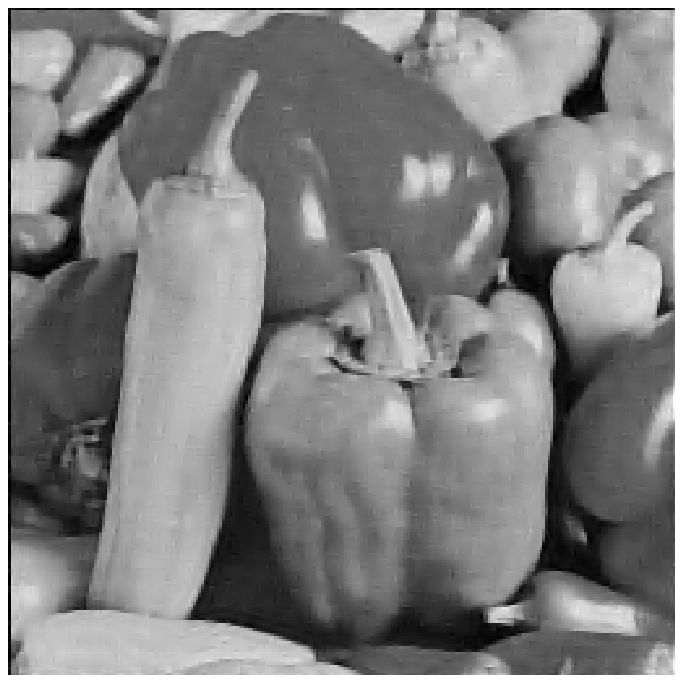}}\\
     \caption{The reconstruction result for {\tt Peppers} for Problem-2, when $K=92$ and $\lambda$ is being tuned based on the different parameter tuning techniques.
     Form top left to bottom right: original image, blurred image, tuning using true ISNR, GCV, L-curve, discrepancy principle, GSURE and projected GSURE.}
     \label{fig:pappers_lambda_exp2}
\end{figure}

Table \ref{tbl:pappers_lambda_exp2} compares the various parameter
setting methods for a deblurring experiment based on Problem-2 for
reconstructing the image {\tt Peppers} when $\lambda$ is being
tuned. The visual effect of the various parameter selections can be observed in Fig.~\ref{fig:pappers_lambda_exp2}.
We see the same behavior as before, indicating that the
projected GSURE is the best choice, leading to the best
reconstruction results. The results again
show that the regular GSURE (and the conventional methods) tend to
fail, while the projected version performs very well.

\begin{table}[htbp]
  \centering
\begin{tabular}{||c|c||}
  \hline
  Parameter Tuning Method & ISNR \\
  \hline
  optimal selection & 8.86dB \\
  projected GSURE & 8.86dB \\
  discrepancy principle & 8.28dB \\
  GSURE & 8.17dB \\
  GCV & 8.1dB \\
  L-curve & 7.51dB \\
  \hline
\end{tabular}
\caption{The deblurring results for Problem-2 applied on {\tt
Peppers} with $\sigma^2=2$, fixing $K = 92$, and setting $\lambda$
using the different parameter tuning techniques.}
\label{tbl:pappers_lambda_exp2}
\end{table}

\begin{figure}[htbp]
     \centering
          \includegraphics[width=.5\textwidth]{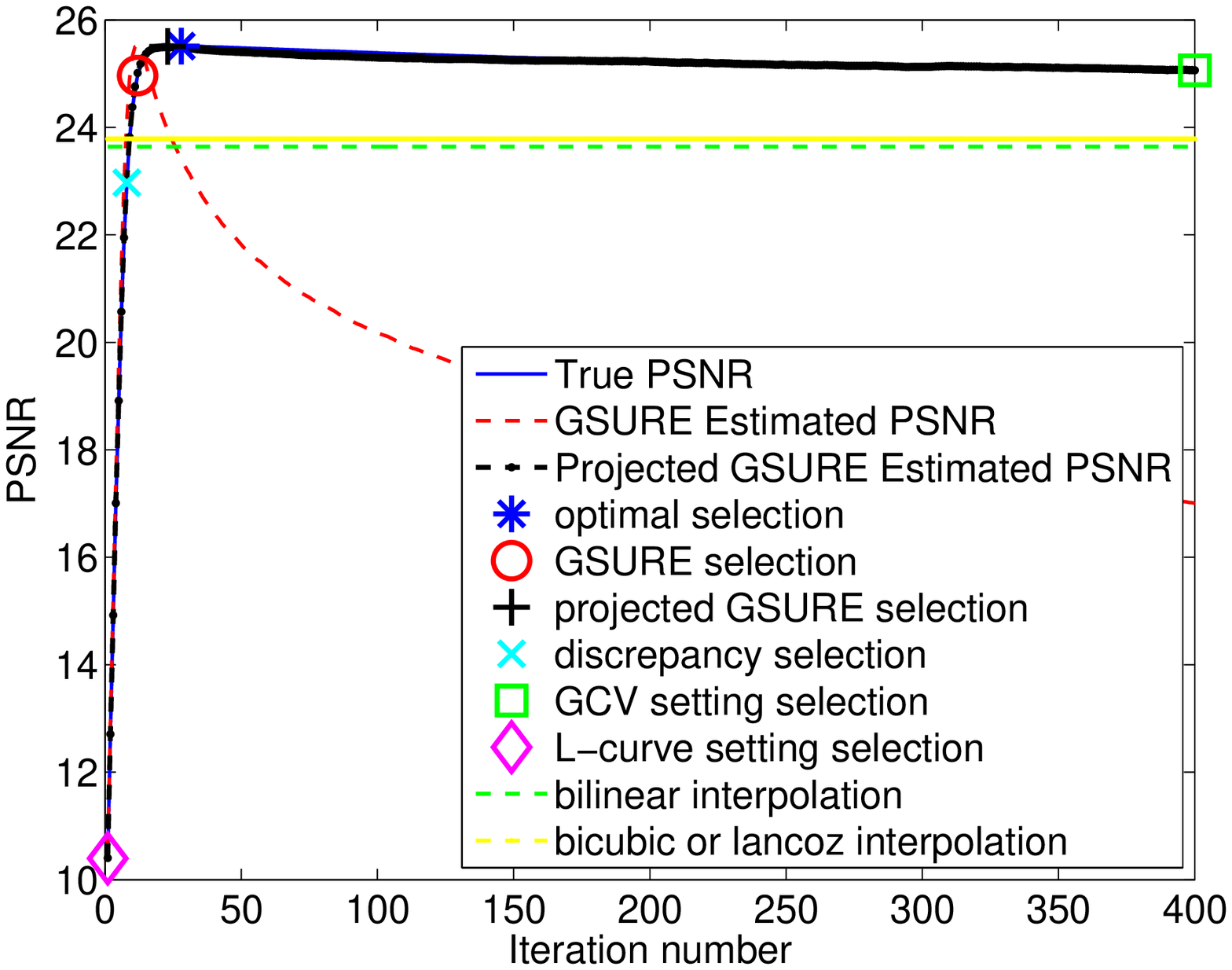}\\
     \caption{The true and estimated PSNR as a function of
     $K$, using the GSURE and its projected version. Applied on Problem-4,
     $\lambda$ is set to be $\lambda = 0.5$, and $K$ is estimated using the various
     proposed methods.}
     \label{fig:iter_GSUREest_upscale}
\end{figure}

\begin{figure}[htbp]
     \centering
          \includegraphics[width=.5\textwidth]{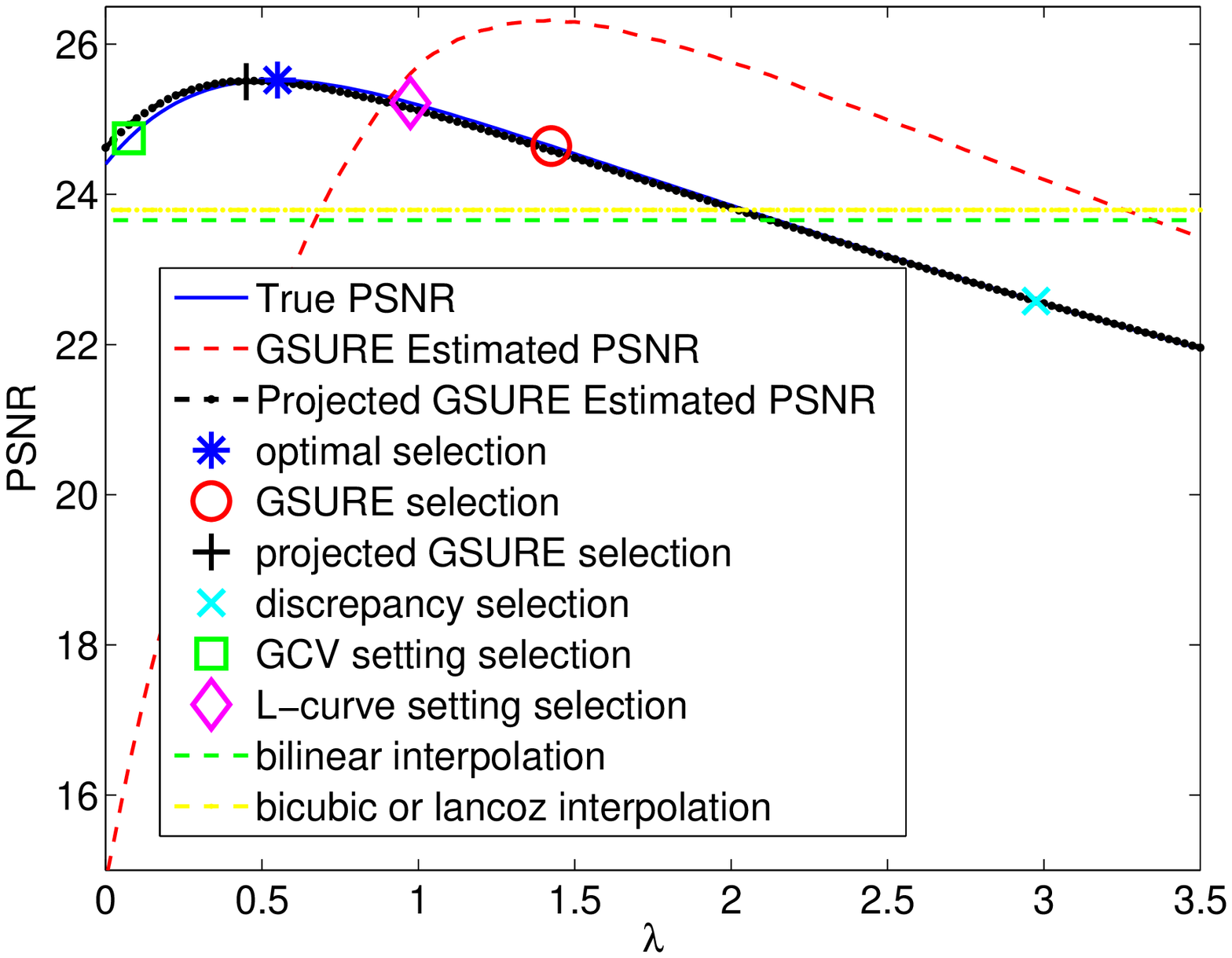}\\
     \vspace{.3in}
     \caption{The true and estimated PSNR as a function of
     $\lambda$, using the GSURE and its projected version. Applied on Problem-4,
     $K$ is set to be $K=27$, and $\lambda$ is estimated using the various
     proposed methods.}
     \label{fig:lambda_GSUREest_upscale}
\end{figure}

In Figs.~\ref{fig:iter_GSUREest_upscale} and
\ref{fig:lambda_GSUREest_upscale} a comparison is made between the
estimation of the PSNR using the regular and the projected GSURE
for Problem 4 (scale-up). As before, a comparison is done between
the parameter values selected by the various methods. In addition,
we show the reconstruction quality achieved by the bicubic, the
Lanczos, and the bilinear interpolation methods\footnote{these scale-up techniques do not require parameter tuning.}, as a reference
performance to compare against.

The first graph is a function of the iterations $K$ and the second
is a function of $\lambda$. It can be seen that the reconstruction
algorithm with correctly tuned parameters achieves better
reconstruction results than the conventional scale-up techniques,
demonstrating the importance of the parameter tuning. Here as
well, it can be observed that the projected GSURE provides the
best estimation for the selection of $\lambda$ and $K$.

As in the case of deblurring, the scale-up results show that the
projected GSURE gives the best reconstruction results. Because of
its superiority, we shall use it hereafter in the various
experiments that follow. Since in the case of a full rank operator,
the GSURE and projected GSURE coincide, in every place that GSURE is
referenced, the projected GSURE would be the one being intended.

\subsection{Advanced Tests}

\begin{figure}[htbp]
     \centering
     \subfloat[greedy.]{
          \includegraphics[width=.48\textwidth]{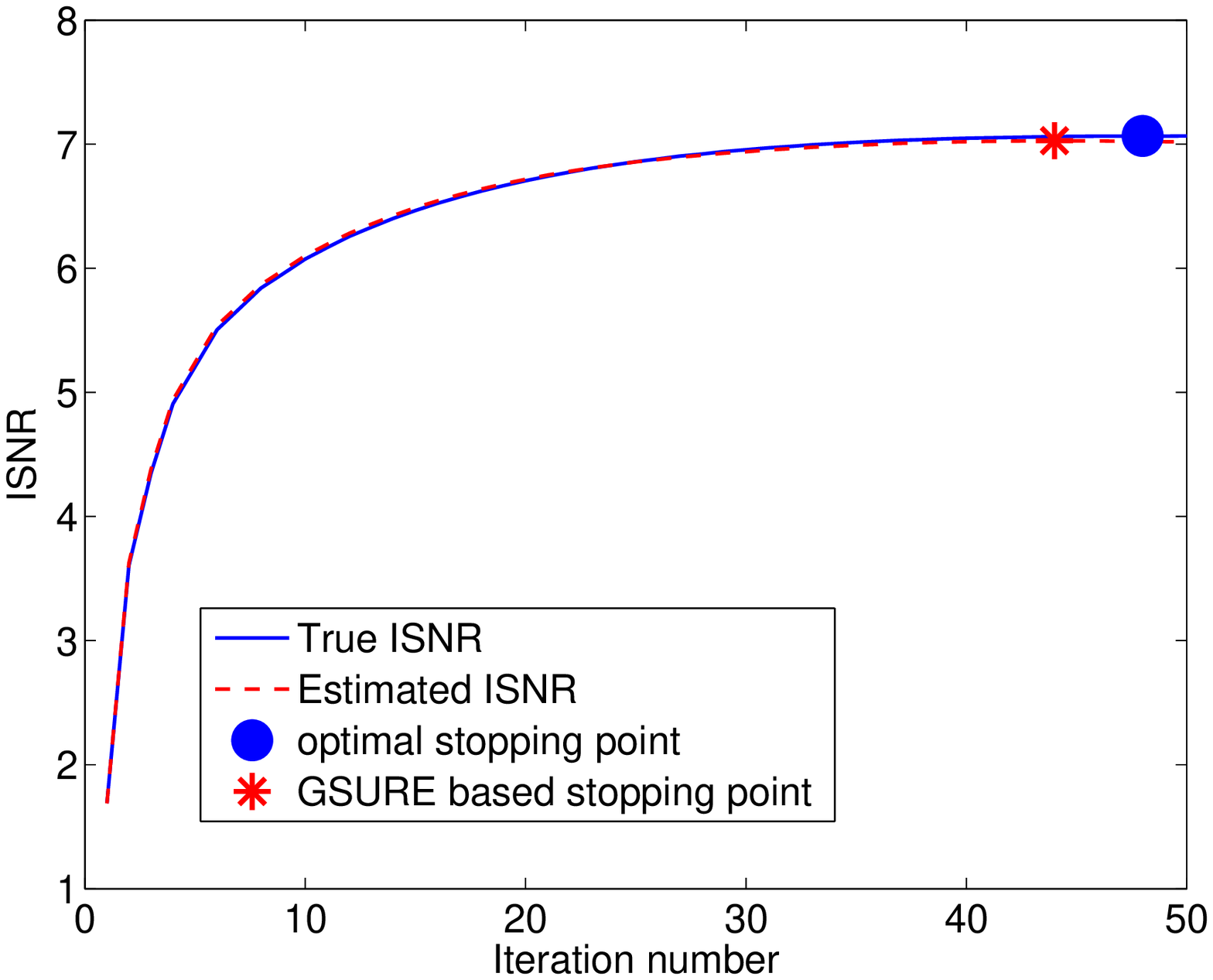}}
     \hspace{.0in}
     \subfloat[greedy look-ahead with $r=1$.]{
          \label{fig:greedyLA2_stopCrit_exp1}
          \includegraphics[width=.48\textwidth]{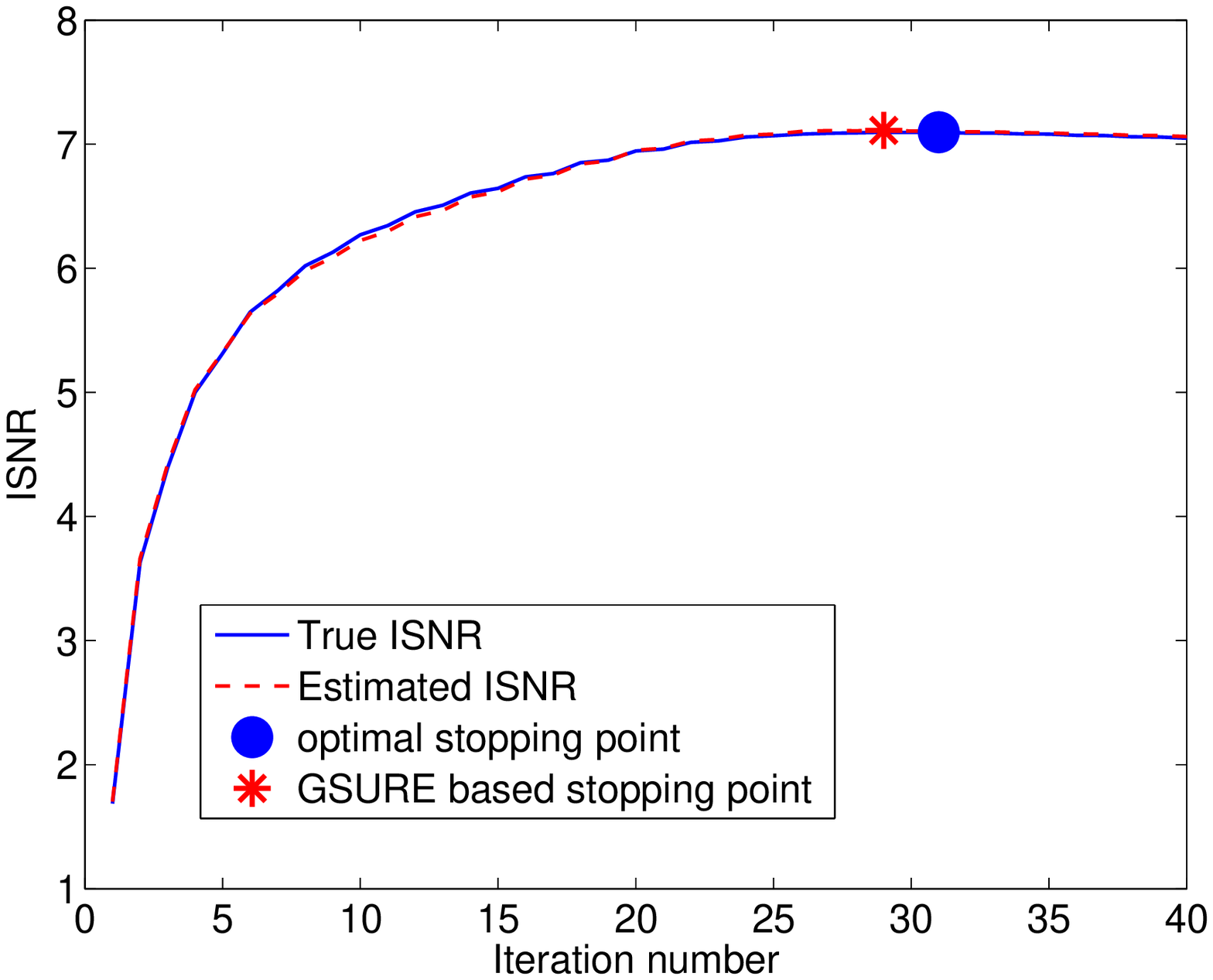}}\\
     \vspace{.0in}
     \caption{The GSURE-estimated and the true ISNR as a function of the
iteration number for the greedy algorithm (top) and for the greedy
look-ahead algorithm with $r=1$ (bottom) for Problem-1.}
     \label{fig:greedy_stopCrit_exp1}
\end{figure}

\begin{figure}[htbp]
     \centering
     \subfloat[original image.]{
          \includegraphics[width=.31\textwidth]{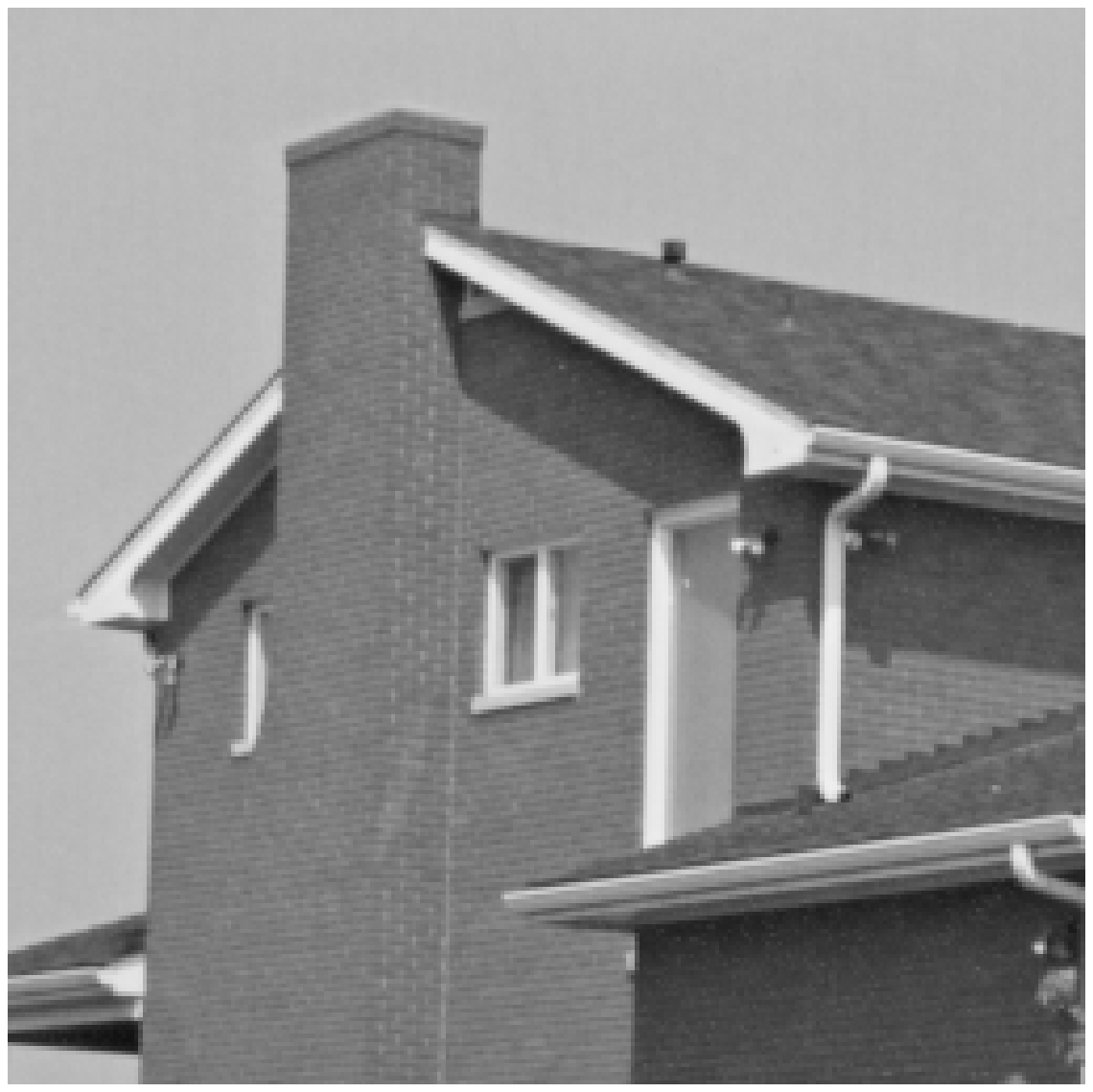}}
     \hspace{.0in}
     \subfloat[blurred image.]{
          \includegraphics[width=.31\textwidth]{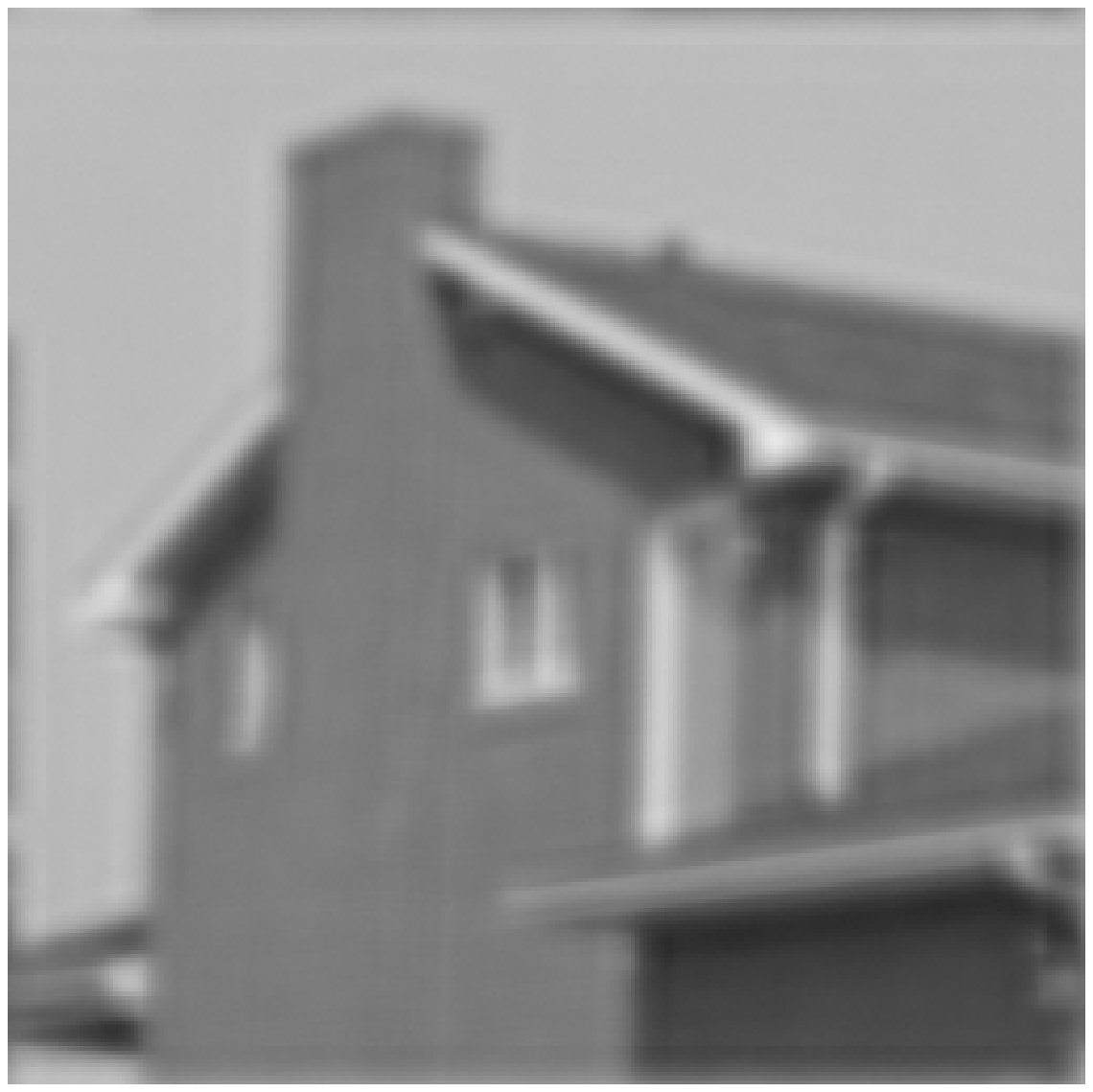}}
     \hspace{.0in}
     \subfloat[greedy look-ahead method with $r=1$. ISNR = 8.65dB.]{
          \includegraphics[width=.31\textwidth]{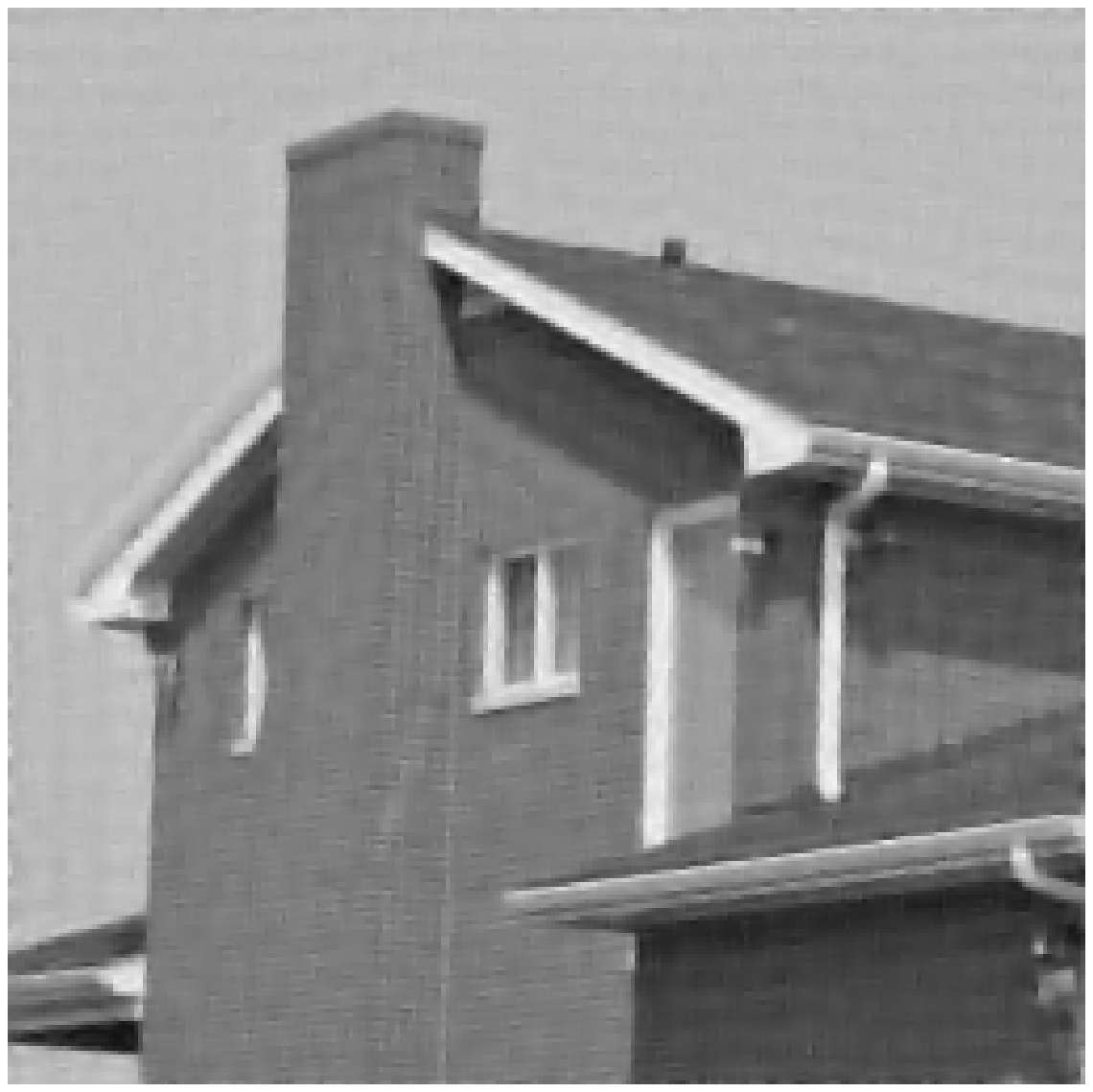}}
     \caption{The reconstruction result for {\tt House} for Problem-2, where $\lambda$ is set based on the
     greedy look-ahead method with $r=1$, and $K$ is set as a stopping rule.}
     \label{fig:house_comp_exp2}
\end{figure}

We now turn to look at the greedy strategies and compare them to
the global method, all aiming to jointly estimate $K$ and
$\lambda$. In Fig.~\ref{fig:greedy_stopCrit_exp1} we present the
results obtained for Problem-1 applied on {\tt Cameraman}. The
graph shows the true ISNR after each iteration of the greedy and
the greedy look-ahead methods (with $r=1$), together with their
ISNR estimation. As can be seen, the estimation has the same
behavior as the true ISNR. In particular, in both graphs the
maxima are very close. Choosing the number of iterations based on
GSURE yields an ISNR very close to the one based on the true ISNR.
This justifies our claim about using the greedy algorithm as an
automatic stopping rule, by detecting an increase in the ISNR. The
maxima in both graphs (marked in a circle as before) are very
close, both in the greedy, as well as in the greedy look-ahead
case. Figure \ref{fig:house_comp_exp2} presents the actual result
obtained for a similar experiment applied on the image {\tt House}
with Problem-2.

\begin{figure}[htbp]
     \centering
     \subfloat[Problem-1 and $\sigma^2=8$.]{
          \label{fig:deblur_AllCompare_ker1_8}
          \includegraphics[width=.33\textwidth]{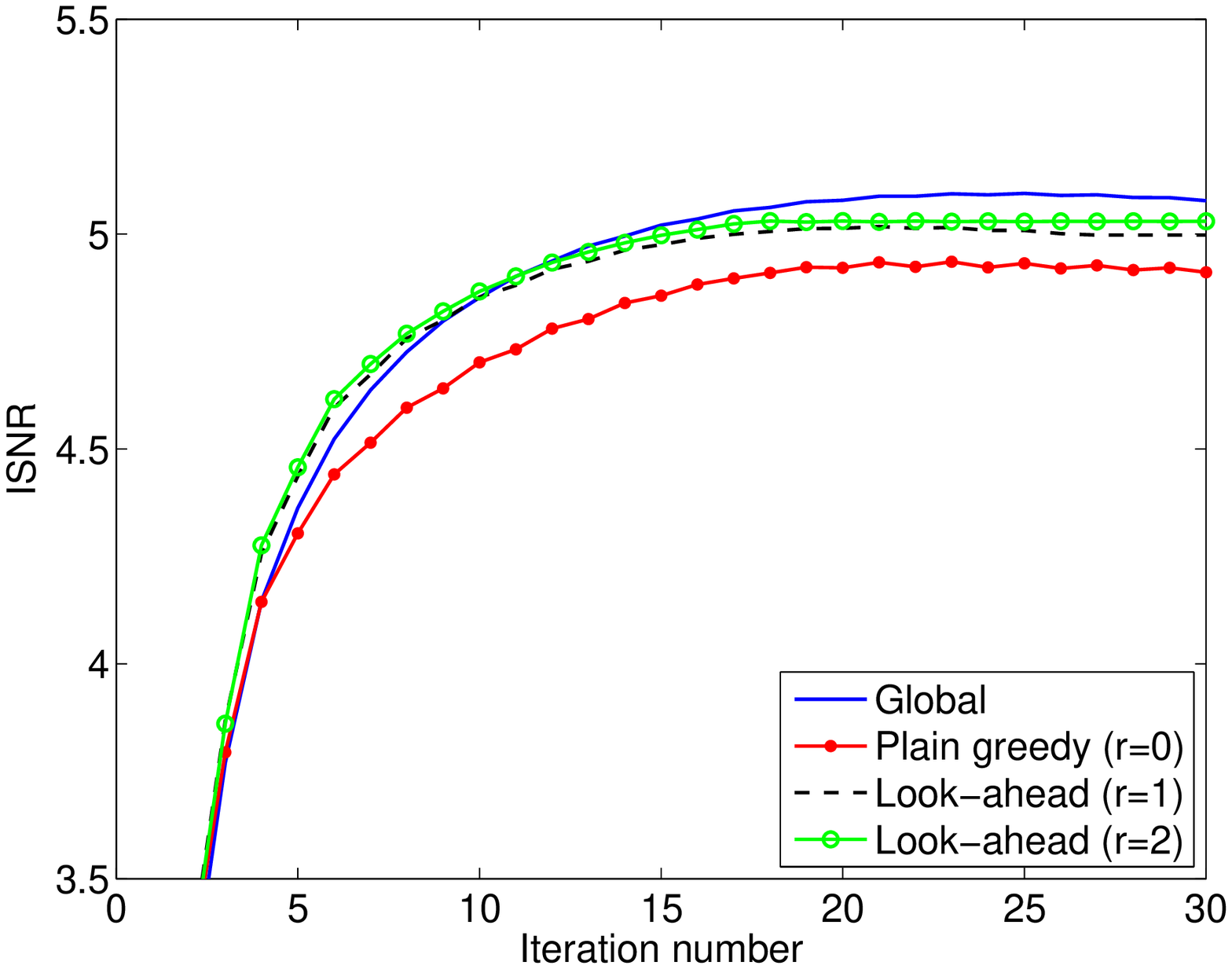}}
     \subfloat[Problem-2.]{
           \label{fig:deblur_AllCompare_ker2}
           \includegraphics[width=.33\textwidth]{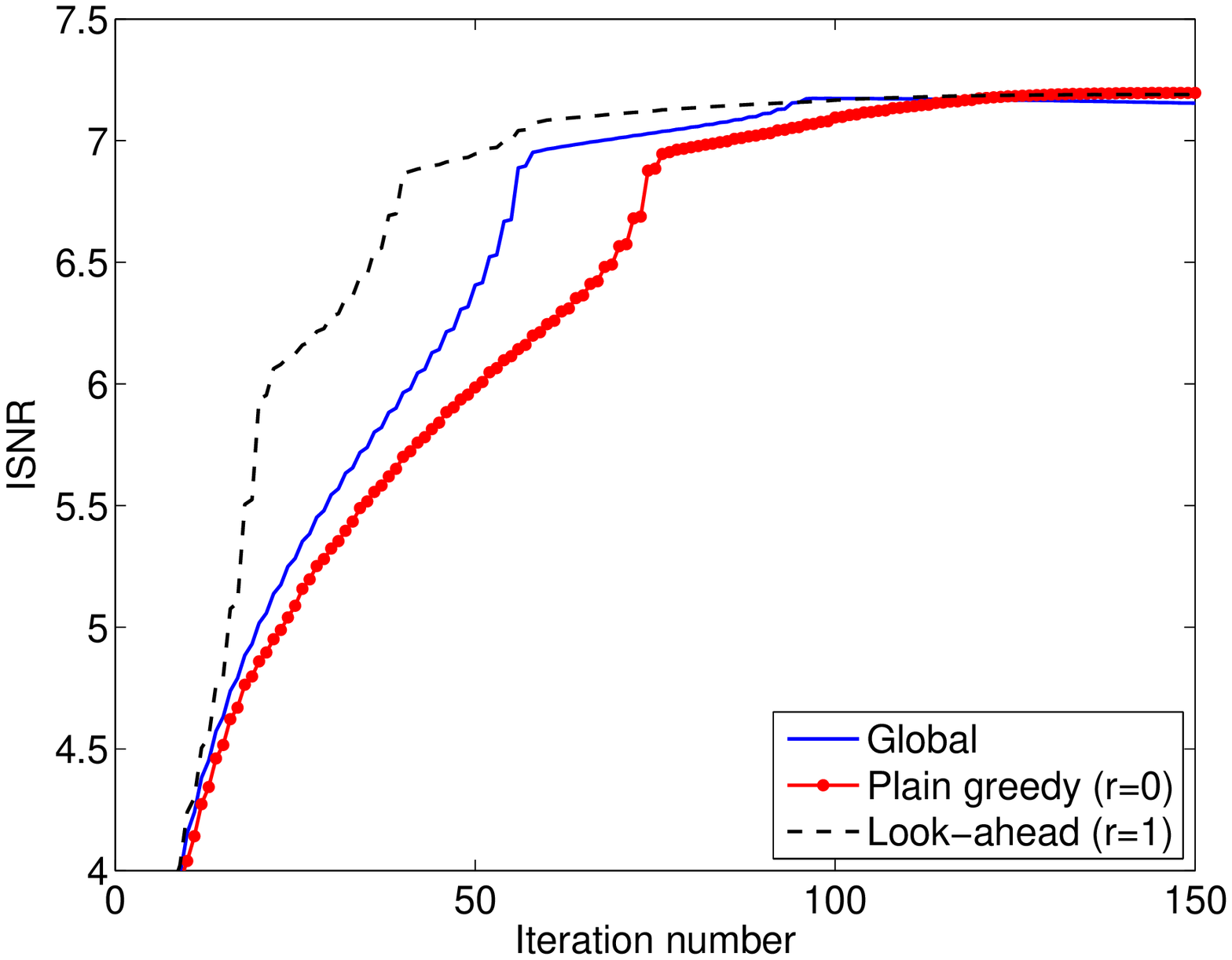}}
     \subfloat[Problem-3.]{
           \label{fig:deblur_AllCompare_ker3}
          \includegraphics[width=.33\textwidth]{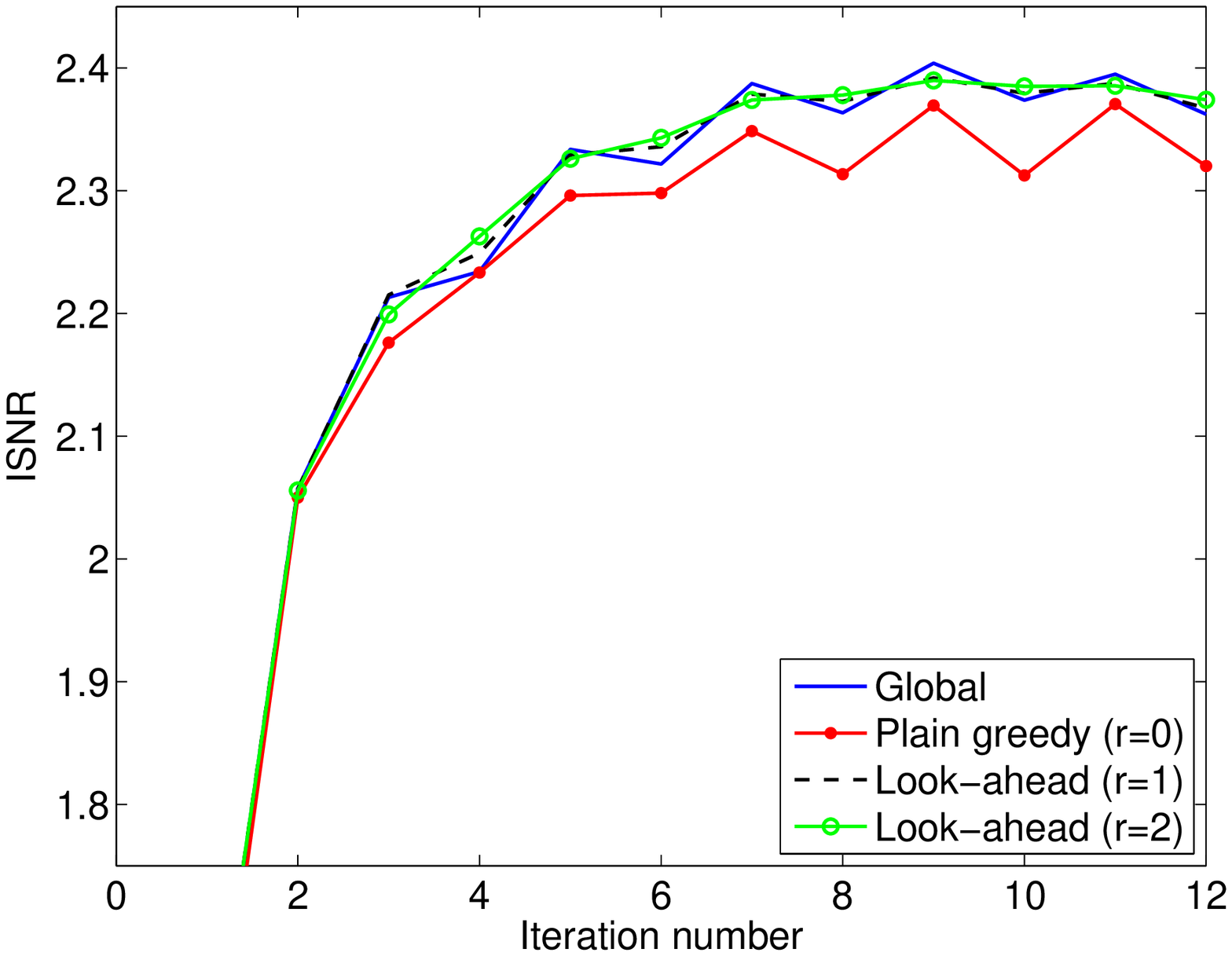}}\\
     \caption{The true ISNR as a function of the iterations number for the
    global, greedy, and look-ahead-greedy methods for Problems 1, 2 and 3 with
    different kernels and noise power.}
     \label{fig:deblur_AllCompare}
\end{figure}

A comparison of the MSE of the global, the greedy, and the
look-ahead techniques for the deblurring problem in images is
presented in Fig.~\ref{fig:deblur_AllCompare} for Problems-1, 2,
and 3 on the image {\tt Cameraman}. In all cases, no real
advantage is seen in terms of ISNR by each of the techniques.
Since the greedy algorithm sets the iterations number together
with $\lambda$, this means that the greedy methods are more
efficient in the overall performance, at a possible slight costs
in terms of the final ISNR.

\begin{figure}[htbp]
     \centering
     \subfloat[Problem-4]{
          \includegraphics[width=.48\textwidth]{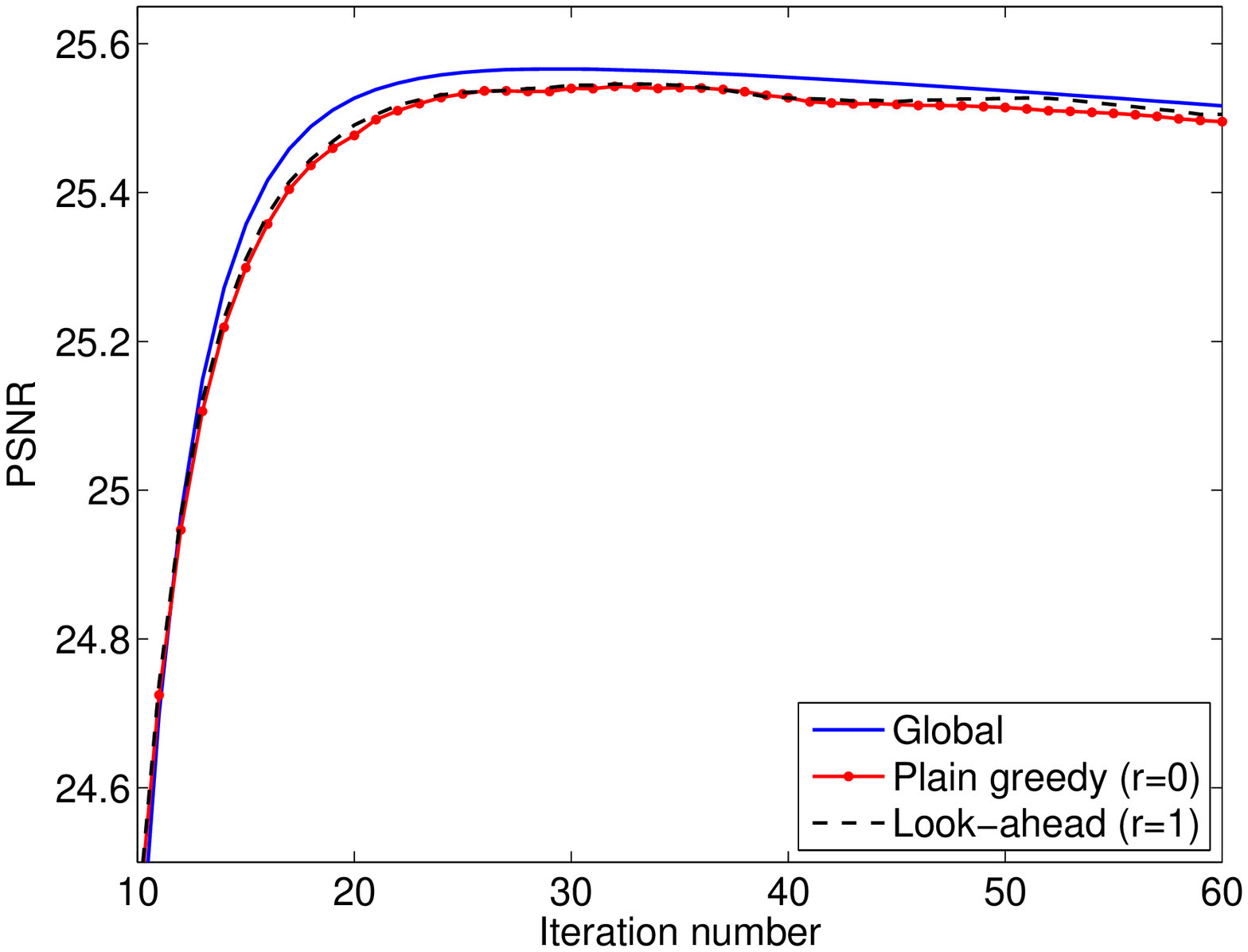}}
     \hspace{.0in}
     \subfloat[Problem-5]{
          \includegraphics[width=.48\textwidth]{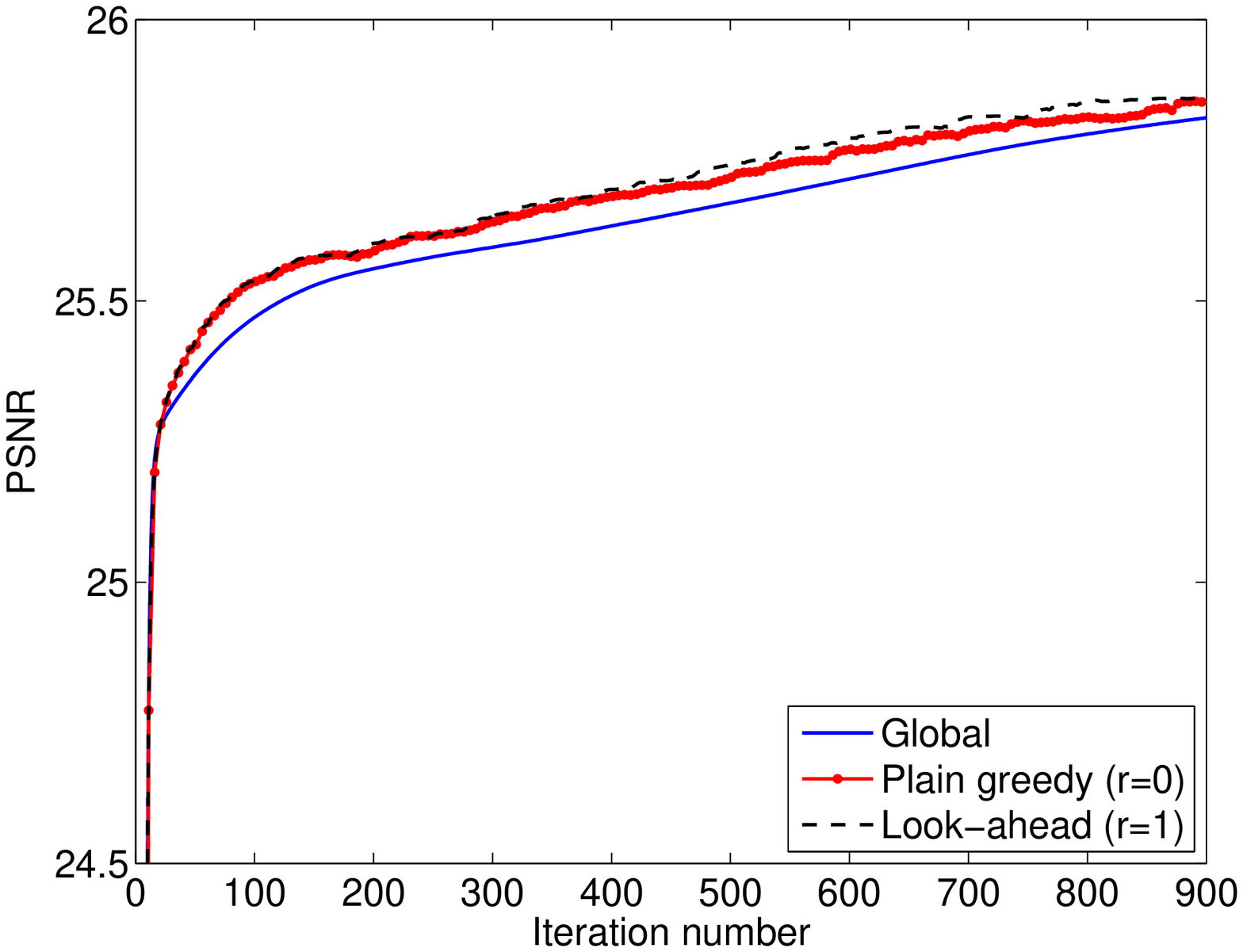}}\\
     \vspace{.0in}
     \caption{The true PSNR as a function of the iterations number for the
    global, greedy, and greedy look-ahead methods for Problems 4 and 5.}
     \label{fig:upscale_AllCompare}
\end{figure}

Turning to the scale-up problems 4 and 5, the same comparison is
done in Fig.~\ref{fig:upscale_AllCompare}. Here again we get
similar results in terms of PSNR for the three methods. This
implies that the two approaches (global and greedy) are equivalent
in terms of the output quality, and the differences are mostly in
their complexities, with a clear advantage to the greedy
techniques.

To summarize, for the various deblurring and scale-up tests, there
is no real advantage to one of the methods over the other, in terms
of the quality of the reconstruction results. Thus, when we need to
set both $\lambda$ and $K$, the greedy methods are to be favored.


\section{Conclusion and Discussion}
\label{sect:conclusion}

In this work we have considered automatic tuning of parameters for
inverse problems. Our focus has been iterated shrinkage algorithms
for deblurring and image scale-up, targeting an automatic way for
choosing $\lambda$ in each iteration in these methods, and
choosing the number of iterations $K$. We extended the global
tuning method developed in \cite{Vonesch08GSURE} to general
ill-posed problems, by exploiting a projected version of the
GSURE. We also applied the GCV, the L-curve, and the
discrepancy-principle methods for this task of parameter tuning,
and compared their selection with those of the (projected) GSURE.
The GSURE was shown to give better approximation for the true
values of the tuned parameters, leading to better results in the
reconstruction.

Two greedy methods for parameter tuning -- a simple and a look-ahead
version -- were presented. These two methods are shown to perform as
well as the global method, with the advantage of setting an
automatic stopping rule together with the $\lambda$ parameter,
whereas the previous methods set only one of the parameters, given
the other.

We are aware of the fact that there are better algorithms today for
image deblurring and scale-up. However, the scope of this paper is
to demonstrate the applicability of the GSURE for parameter tuning
for such methods and not to compete with current state of the art
reconstruction results. Using the concepts presented in this paper,
automatic parameter tuning can be performed also for other
techniques.

\section*{Acknowledgment}
The authors are grateful to Mario A.T. Figueiredo for sharing his image
restoration software with them.

This research was partly supported by the European Community's FP7-FET
program, SMALL project, under grant agreement no. 225913, by the ISF
grant number 599/08 and by the
Israel Science Foundation under Grant no. 1081/07 and by the
European Commission in the framework of the FP7 Network of
Excellence in Wireless COMmunications NEWCOM++ (contract no.
216715).

\bibliographystyle{elsarticle-num}
\bibliography{AutomaticTuning}

\end{document}